\documentclass{article}

\PassOptionsToPackage{numbers, compress}{natbib}
\usepackage[final]{neurips_2022}
\usepackage[utf8]{inputenc} 
\usepackage[T1]{fontenc}    
\usepackage{hyperref}       
\usepackage{url}            
\usepackage{booktabs}       
\usepackage{amsfonts}       
\usepackage{nicefrac}       
\usepackage{microtype}      
\usepackage{xcolor}         
\usepackage{graphicx}
\usepackage{caption}
\usepackage{subfig}
\usepackage{wrapfig}
\usepackage{multirow}
\usepackage[noend]{algpseudocode}
\usepackage{algorithmicx,algorithm}

\title{\textsc{AD-Drop}: Attribution-Driven Dropout for Robust Language Model Fine-Tuning}

\author{%
  Tao Yang\textsuperscript{1}, Jinghao Deng\textsuperscript{1}, Xiaojun Quan\textsuperscript{1}\thanks{Corresponding author.}, Qifan Wang\textsuperscript{2}, Shaoliang Nie\textsuperscript{2}\\
  \textsuperscript{1}School of Computer Science and Engineering, Sun Yat-sen University \textsuperscript{2}Meta AI\\
  \textsuperscript{1}\texttt{\{yangt225,dengjh27\}@mail2.sysu.edu.cn, quanxj3@mail.sysu.edu.cn} \\
  \textsuperscript{2}\texttt{\{wqfcr, snie\}@fb.com} \\
}
  
\begin{document}

\maketitle

\begin{abstract}
Fine-tuning large pre-trained language models on downstream tasks is apt to suffer from overfitting when limited training data is available. While dropout proves to be an effective antidote by randomly dropping a proportion of units, existing research has not examined its effect on the self-attention mechanism. In this paper, we investigate this problem through self-attention attribution and find that dropping attention positions with low attribution scores can accelerate training and increase the risk of overfitting. Motivated by this observation, we propose Attribution-Driven Dropout (\textsc{AD-Drop}), which randomly discards some high-attribution positions to encourage the model to make predictions by relying more on low-attribution positions to reduce overfitting. We also develop a cross-tuning strategy to alternate fine-tuning and \textsc{AD-Drop} to avoid dropping high-attribution positions excessively. Extensive experiments on various benchmarks show that \textsc{AD-Drop} yields consistent improvements over baselines.~Analysis further confirms that \textsc{AD-Drop} serves as a strategic regularizer to prevent overfitting during fine-tuning.

\end{abstract}

\section{Introduction}
\label{intr}
Pre-training large language models (PrLMs) on massive unlabeled corpora and fine-tuning them on downstream tasks has become a new paradigm \cite{devlin2018bert,liu2019roberta,brown2020language}. Their success can be partly  attributed to the self-attention mechanism \cite{vaswani2017attention}, yet these self-attention networks are often redundant 
\cite{ZaheerGDAAOPRWY20,JaszczurCMKGMK21} and tend to cause overfitting when fine-tuned on downstream tasks due to the mismatch between their overparameterization and the limited annotated data \cite{michel2019sixteen, kovaleva2019revealing,clark2019does, dong2021attention,lee2019mixout,mosbach2020stability,xu2021raise}. To address this issue, various regularization techniques such as data augmentation \cite{andreas2020good,chen2021hiddencut}, adversarial training \cite{liu2020adversarial,miao2021simple}), and dropout-based methods \cite{lee2019mixout,xu2021raise,wu2021r} have been developed. Among them, dropout-based methods are widely adopted for their simplicity and effectiveness. Dropout \cite{srivastava2014dropout}, which randomly discards a proportion of units, is at the core of dropout-based methods. Recently, several variants of dropout have been proposed, such as Concrete Dropout \cite{gal2017concrete}, DropBlock \cite{ghiasi2018dropblock}, and AutoDropout \cite{pham2021autodropout}. However, these variants generally follow the vanilla dropout to randomly drop units during training and pay little attention to the effect of dropout on self-attention. In this paper, we seek to fill this gap from the perspective of self-attention attribution \cite{hao2021self} and aim to reduce overfitting when fine-tuning PrLMs.

Attribution \cite{zhang2021survey} is an interpretability method that attributes model predictions to input features via saliency measures such as gradient \cite{baehrens2010explain,sundararajan2017axiomatic}. It is also used to explain the influence patterns of self-attention in recent literature \cite{hao2021self,jain2019attention,lu2021influence}. Our prior experiment of self-attention attribution (Section \ref{prior}) reveals that attention positions are not equally important in preventing overfitting, and dropping low-attribution positions is more likely to cause overfitting than discarding high-attribution positions. This observation suggests that attention positions should not be treated the same in dropout.

\begin{wrapfigure}{r}{7.4cm}
 	\centering
 		\includegraphics[width=0.51\textwidth]{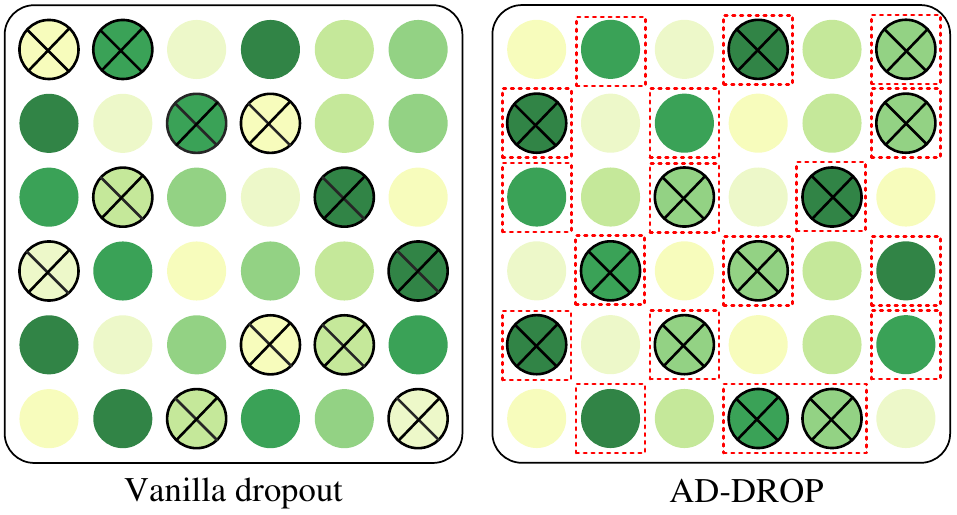}
 		\vspace{-0.1cm}
 	\caption{Attention maps of vanilla dropout and our \textsc{AD-Drop}. Darker attention positions indicate higher attribution scores, and crossed circles mean dropped attention positions. Red-dotted boxes refer to candidate discard regions with high attribution scores. Unlike vanilla dropout which randomly discards attention positions, \textsc{AD-Drop} focuses on dropping high-attribution positions in candidate discard regions.}\vspace{-0.2cm}
 	\label{fig:attention_maps}
\end{wrapfigure}

Motivated by the above, we propose \textbf{A}ttribution-\textbf{D}riven \textbf{Drop}out (\textsc{AD-Drop}) to better fine-tune PrLMs based on self-attention attribution. The general idea of \textsc{AD-Drop} is to drop a set of self-attention positions with high attribution scores. We illustrate the difference between vanilla dropout and \textsc{AD-Drop} by their attention maps in Figure \ref{fig:attention_maps}.~When fine-tuning a PrLM on a batch of training samples, \textsc{AD-Drop} involves four steps. First, predictions are made through a forward computation without dropping any attention position. Second, we compute the attribution score of each attention position by gradient \cite{baehrens2010explain} or integrated gradient \cite{sundararajan2017axiomatic} attribution methods. Third, we sample a set of positions with high attribution scores and generate a mask matrix for each attention map. Finally, the mask matrices are applied to the next forward computation to make predictions for backpropagation. \textsc{AD-Drop} can be regarded as a strategic dropout regularizer that forces the model to make predictions by relying more on low-attribution positions to reduce overfitting. Nevertheless, excessive neglect of high-attribution positions would leave insufficient information for training. Hence, we further propose a cross-tuning strategy that performs fine-tuning and \textsc{AD-Drop} alternately to improve the training stability. 

To verify the effectiveness of \textsc{AD-Drop}, we conduct extensive experiments with different PrLMs (i.e., BERT \cite{devlin2018bert}, RoBERTa \cite{liu2019roberta}, ELECTRA \cite{clark2020electra}, and OPUS-MT \cite{TiedemannThottingal:EAMT2020}) on various datasets (i.e., GLUE \cite{wang2018glue}, CoNLL-2003 \cite{sang2003introduction}, WMT 2016 EN-RO and TR-EN \cite{bojar2016findings}, HANS \cite{mccoy2019right}, and PAWS-X \cite{yang2019paws}). Experimental results show that the models tuned with \textsc{AD-Drop} obtain remarkable improvements over that tuned with the original fine-tuning approach. For example, on the GLUE benchmark, BERT achieves an average improvement of 1.98/0.87 points on the dev/test sets while RoBERTa achieves an average improvement of 1.29/0.62 points. Moreover, ablation studies and analysis demonstrate that gradient-based attribution \cite{baehrens2010explain,sundararajan2017axiomatic} is a more suitable saliency measure for implementing \textsc{AD-Drop} than directly using attention weights or simple random sampling. Moreover, they also demonstrate that the cross-tuning strategy plays a crucial role in improving training stability.

To sum up, this work reveals that self-attention positions are not equally important for dropout when fine-tuning PrLMs. Arguably, low-attribution positions are more difficult to optimize than high-attribution positions, and dropping these positions tends not to relieve but accelerate overfitting. This leads to a novel dropout regularizer, \textsc{AD-Drop}, driven by self-attention attribution. Although proposed for self-attention units, \textsc{AD-Drop} can be potentially extended to other units as dropout.

\section{Methodology}

\subsection{Preliminaries}
\label{preliminaries}
Since Transformers \cite{vaswani2017attention} are the backbone of PrLMs, we first review the details of self-attention in Transformers and self-attention attribution \cite{hao2021self}. Let ${\bf{X}} \in {\mathbb{R}^{n \times d}}$ be the input of a Transformer block, where $n$ is the sequence length and $d$ is the embedding size. Self-attention in this block first maps $\bf{X}$ into three matrices ${\bf{Q}}_h$, ${\bf{K}}_h$ and ${\bf{V}}_h$ via linear projections as query, key, and value respectively for the $h$-th head. Then, the attention output of this head is calculated as:
\begin{equation} \label{eq2}
{\rm{Attention}}\left( {{\bf{Q}}_h,{\bf{K}}_h,{\bf{V}}_h} \right) = {{\bf{A}}_h}{{\bf{V}}_h} = {\rm{softmax}}\left( {\frac{{{\bf{Q}}_h{{{\bf{K}}_h}^{\rm{T}}}}}{{\sqrt {d_k} }}} + {\bf{M}}_h \right){\bf{V}}_h, 
\end{equation}
where $\sqrt{d_k}$ is a scaling factor. ${\bf{M}}_h$ is the mask matrix to apply dropout in self-attention, and elements in ${\bf{M}}_h$ will be $- \infty$ if the corresponding positions in attention maps are masked and 0 otherwise. 

Based on the attention maps ${\bf{A}} = \left[ {{\bf{A}}_1,{\bf{A}}_2, \cdots ,{\bf{A}}_H} \right]$ for $H$ attention heads, gradient attribution \cite{baehrens2010explain, simonyan2014deep} directly produces an attribution matrix $\bf{B}_h$ by computing the following partial derivative:
\begin{equation} \label{eq4}
{\bf{B}}_h =  {\frac{\partial {F}_{c} \left( {\bf{A}} \right)}{{\partial {\bf{A}}_h}}},
\end{equation}
where $F_{{c}} {\left( {\cdot} \right)}$ denotes the logit output of the Transformer for class $c$.

To provide a theoretically more sound attribution method, \citet{sundararajan2017axiomatic} propose integrated gradient, which is employed by \citet{hao2021self} as a saliency measure for self-attention attribution. Specifically, \citet{hao2021self} compute the attribution matrix ${\bf{B}}_h$ as:
\begin{equation} \label{eq3}
{{{{\bf{B}}}}_h} = \frac{{{\bf{A}}_h}}{m} \odot \sum\limits_{k = 1}^m {\frac{{\partial {F}_{c} \left( {\frac{k}{m}{\bf{A}}} \right)}}{{\partial {\bf{A}}_h}}},
\end{equation}
where $m$ is the number of steps for approximating the integration in integrated gradient, and $\odot$ is the element-wise multiplication operator.~Despite its theoretical advantage over gradient attribution, integrated gradient requires $m$ times more computational effort, which is especially expensive when it is applied to all the attention heads in Transformers.~Moreover, our experiments in Section \ref{ablation} show that gradient attribution achieves comparable performance with integrated gradient but requires much less computational cost, suggesting that gradient attribution is more desirable for \textsc{AD-Drop}.

\subsection{A Prior Attribution Experiment}
\label{prior}
\begin{wrapfigure}{r}{8.5cm}
	\centering
	 \vspace{-0.4cm}
	\includegraphics[width=0.6\textwidth]{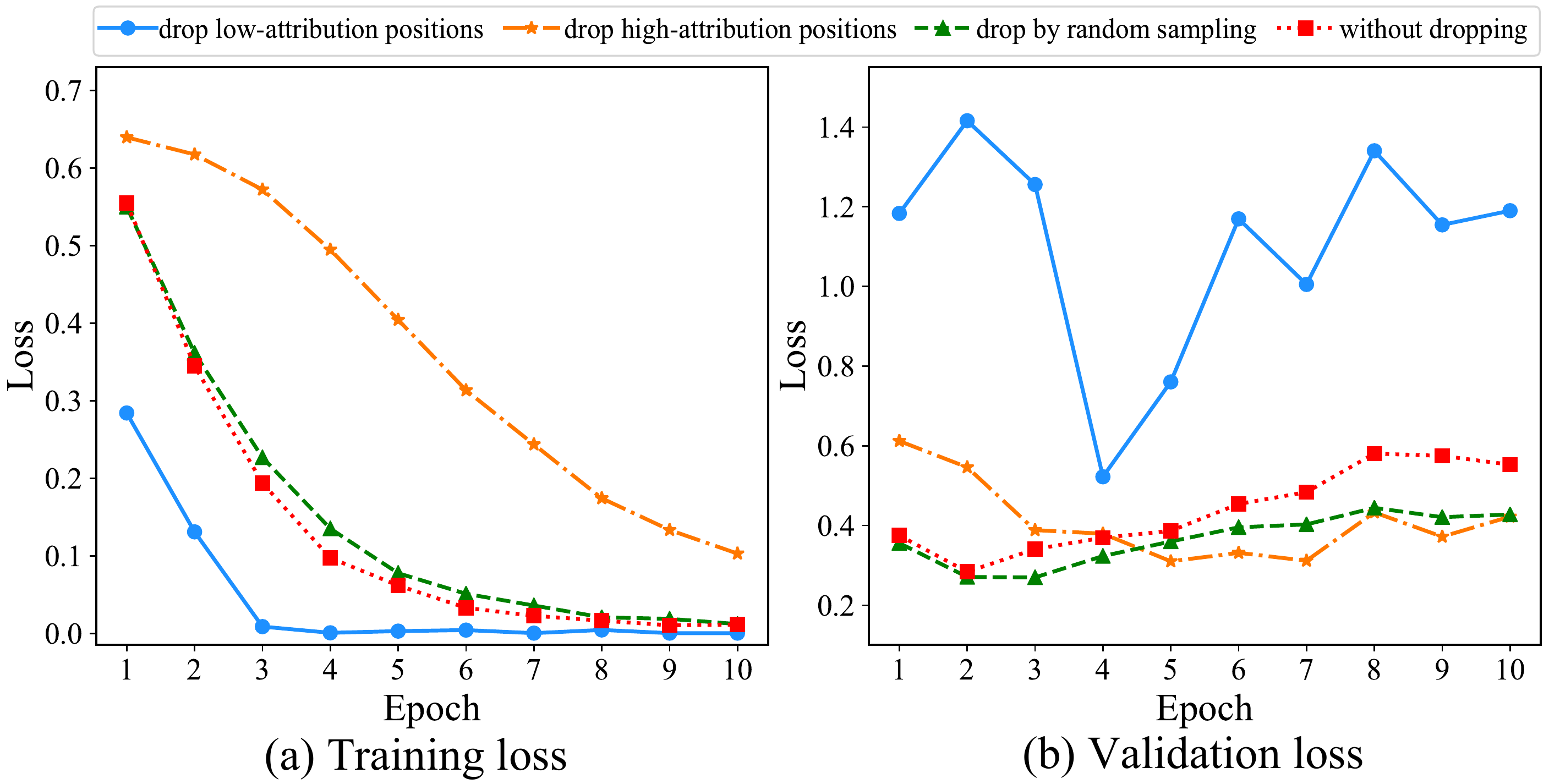}
	\vspace{-0.1cm}
	\caption{Results of training and validation losses when fine-tuning RoBERTa with different dropping strategies on MRPC. The dropping rate is set to 0.3 if it applies.}
	\label{fig:prior_results}
	 	\vspace{-0.3cm}
\end{wrapfigure}

To better motivate our work, we first conduct a prior experiment on MRPC \cite{dolan2005automatically} to investigate how different positions in self-attention maps affect fine-tuning performance based on attribution results. \texttt{RoBERTa\_base} \cite{liu2019roberta} is used as the base model.~To begin with, we first perform a forward computation of the model on each batch of training samples to obtain the logit output of each sample corresponding to the gold label. Then, we obtain an attribution matrix ${\bf{B}}_h$ for the self-attention positions in the first layer\footnote{We provide more results and discussions in Appendix \ref{apd:probe}.} by gradient attribution with Eq.~(\ref{eq4}) and sort each row of the matrix. Finally, we sample a set of self-attention positions with high or low attribution scores in each row to generate a mask matrix ${\bf{M}}_h$, which is fed into Eq.~(\ref{eq2}) to make the final predictions. After each epoch of training, we evaluate the model on the development set. Two baseline dropping strategies (i.e., dropping by random sampling and without dropping any position) are employed for comparison. We plot the loss curves of the model with these dropping strategies on both training and development sets in Figure \ref{fig:prior_results}. The observations are threefold. First, dropping low-attribution positions makes the model fit the training data rapidly, whereas it performs poorly on the development set, indicating that the model is not properly trained. Second, compared with the other dropping strategies, dropping high-attribution positions reduces the fitting speed significantly. Third, random dropping only slightly reduces overfitting, compared to the training without dropping. These observations suggest that attention positions are of different importance in preventing overfitting. We conjecture that low-attribution positions are more difficult to optimize than high-attribution positions. While dropping low-attribution positions tends to accelerate overfitting, discarding high-attribution positions helps reduce overfitting. 

\subsection{Attribution-Driven Dropout}
\label{addrop}
Inspired by the observations in Section \ref{prior}, we propose a novel regularizer, \textsc{AD-Drop}, to better prevent overfitting when adapting PrLMs to downstream tasks. The motivation of \textsc{AD-Drop} is to minimize the over-reliance of these models on particular features which may affect their generalization. Formally, given a training set $\mathcal{D} = \left\{ {\left( {{{{x}}_i},{{{y}}_i}} \right)} \right\}_{i = 1}^N$ of $N$ samples, where $x_i$ is the $i$-th sample and $y_i$ is its label, the goal of \textsc{AD-Drop} is to fine-tune a PrLM $F{\left( {\cdot} \right)}$ of $L$ layers on $\mathcal{D}$. Same as the vanilla dropout \cite{srivastava2014dropout}, \textsc{AD-Drop} is only applied in the training phase.

\begin{figure*}[htbp]
	\centering
	\includegraphics[width=1.0\textwidth]{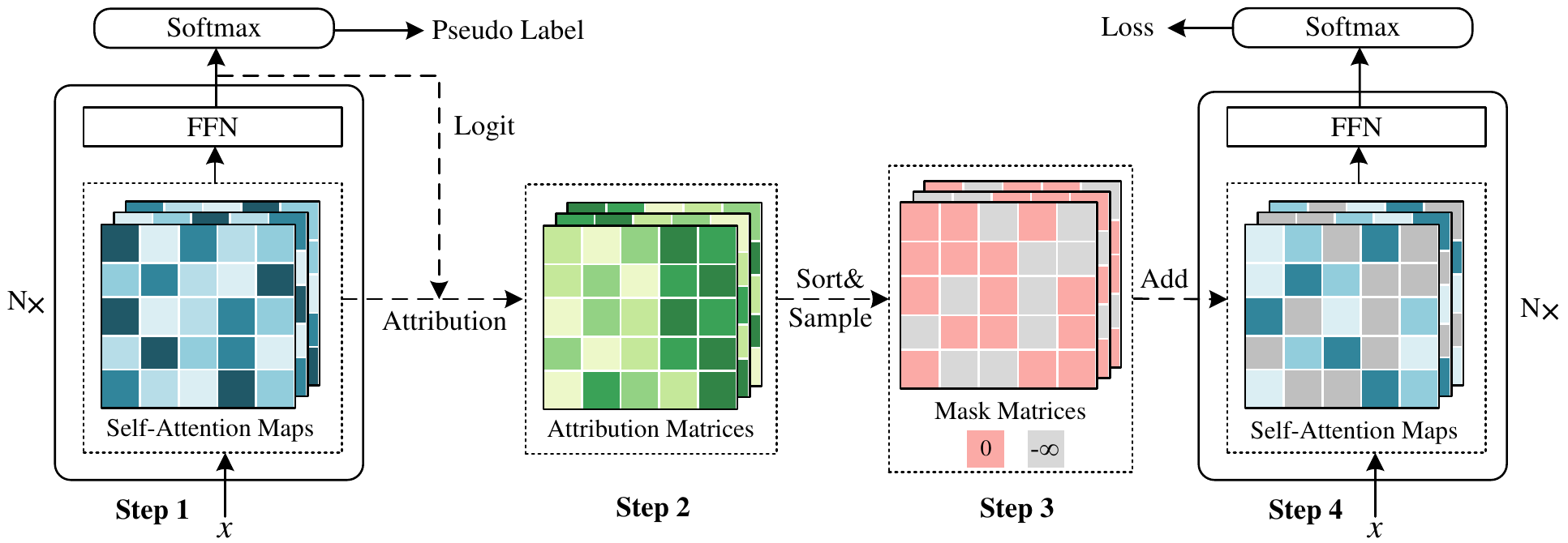}
	\caption{Illustration of \textsc{AD-Drop} in four steps. (1) Conduct the first forward computation to obtain pseudo label $\tilde {c}$. (2) Generate attribution matrices ${\bf{B}}$ via computing the gradient of logit output ${F_{\tilde {c}}}\left( {\bf{A}} \right)$ with respect to each attention head. (3) Sort ${\bf{B}}$ and strategically drop some positions to produce mask matrices ${\bf{M}}$. (4) Feed ${\bf{M}}$ into the next forward computation to compute the final loss.}
	\label{fig:method}
\end{figure*}

As shown in Figure \ref{fig:method}, the idea of \textsc{AD-Drop} can be described in four steps. First, we conduct a forward computation of the model to obtain the label with the highest probability as the pseudo label. The reason we adopt pseudo labels rather than gold labels for attribution will be explained shortly. Specifically, for the input $x_i$ with $n$ tokens, we apply $F{\left( {\cdot} \right)}$ to encode it and obtain its pseudo label $\tilde {c}$:
\begin{equation} \label{eq5}
\tilde {{c}} = {\mathop{\arg\max}_{c}}\left( {P_F}\!\left( {{c}|{x_i}} \right) \right),
\end{equation}
where ${P_F}\!\left( c |x_i \right)$ is the probability of class $c$ for $x_i$. After the forward computation, we also obtain a set of attention maps ${\bf{A}} = \left[ {{\bf{A}}_1,{\bf{A}}_2, \cdots ,{\bf{A}}_H} \right]$ for each layer according to Eq.~(\ref{eq2}). 

Second, we compute the attribution matrices ${\bf{B}} = \left[ {{\bf{B}}_1,{\bf{B}}_2, \cdots ,{\bf{B}}_H} \right]$ for $H$ heads according to Eq.~(\ref{eq4}). Specifically, the attribution matrix ${\bf{B}}_h$ for the $h$-th head is computed as:
\begin{equation} \label{eq:attribution_matrix}
{{\bf{B}}_h = {\frac{{\partial F_{\tilde {{c}}} \left( {{\bf{A}}} \right)}}{{\partial {\bf{A}}_h}}}},
\end{equation}
where $F_{\tilde c} \left( {{\bf{A}}} \right)$ is the logit output of pseudo label $\tilde c$ before softmax.\footnote{The negative loss will be used for both regression and token-level tasks, as introduced in Appendix \ref{apd:tok}.}

Third, we generate a mask matrix ${\bf{M}}_h$ based on ${\bf{B}}_h$. To this end, we first sort each row of ${\bf{B}}_h$ in ascending order and obtain a sorted attribution matrix ${{\bf{\widehat B}}_h}$. Then, we define a candidate discard region ${\bf{S}}_h$, in which each element $s_{i,j}$ is defined as:
\begin{equation} \label{eq:discard_region}
s_{i,j} = \left\{ {\begin{array}{*{20}{c}}
	1,&b_{i,j} < \widehat{b}_{i,\mathop{\rm int} \left(n \left(1 - p \right) \right)}\\
	0,&{{\rm{otherwise}}}
	\end{array}} \right.
\end{equation}
where $b_{i,j}$ and $\widehat {b}_{i,j}$ are elements of ${\bf{B}}_h$ and ${{\bf{\widehat B}}_h}$, respectively, ${\rm int}\!\left(\cdot\right)$ is an integer function, and $p \in \left( {0,1} \right)$ is used to control the size of the candidate discard region. Next, we apply dropout in the region to produce the mask matrix ${\bf{M}}_h$ as:
\begin{equation} \label{eq8}
m_{i,j} = \left\{ {\begin{array}{*{20}{c}}
	{ - \infty ,}&{\left( {{{{s}}_{i,j}}{\rm{ + }}{u_{i,j}}} \right) = 0}\\
	0,&{{\rm{otherwise}}}
	\end{array}} \right.
\end{equation}
where $u_{i,j} \sim {{\rm Bernoulli}\!\left(1 - q \right)}$ is an element of matrix ${{\bf{U}}_h} \in {\mathbb{R}^{n \times n}}$, and $q$ is the dropout rate.

Finally, ${\bf{M}}_h$ is fed into self-attention of Eq.~(\ref{eq2}) for the second forward computation, and the final output is used to calculate the loss for backpropagation.

\paragraph{Discussion}The reasons that \textsc{AD-Drop} uses pseudo labels for attribution are twofold. First, adopting gold labels will divulge label information and lead to inconsistency between training and inference. Second, for misclassified samples in the first forward computation, \textsc{AD-Drop} with gold labels tends to continue to make incorrect predictions because high-attribution attention positions derived from gold labels may be located in low-attribution regions derived from pseudo labels. Therefore, dropping these positions does not help the model correct wrong predictions, while \textsc{AD-Drop} with pseudo labels urges the model to rely on important features in the current pass and may correct the wrong predictions. The attribution with gold labels will be investigated in Section \ref{ablation}.

\subsection{Cross-Tuning Algorithm}
We further design a cross-tuning algorithm to avoid dropping high-attribution positions excessively when applying \textsc{AD-Drop}. The idea of cross-tuning is to execute the original fine-tuning and \textsc{AD-Drop} alternatively. Specifically, it performs the original fine-tuning at odd epochs and \textsc{AD-Drop} at even epochs. The overall process of cross-tuning is described in Algorithm \ref{alg:cross}, where Lines 3-5 are the original fine-tuning operations and Lines 7-9 describe the process of \textsc{AD-Drop}.

\begin{algorithm}[htbp]
	\caption{Cross-tuning} 
	\hspace*{0.02in} {\bf Input:} 
	shuffled training samples $\mathcal{D} = \left\{ {\left( {{x_i},{y_i}} \right)} \right\}_{i = 1}^N$, PrLM $F$ with parameters $\bf W$\\
	\hspace*{0.02in} {\bf Output:} 
	updated parameters ${\bf {\widetilde {W}}}$
	\begin{algorithmic}[1]
		\State Initialize $F$ with $\bf W$, $epoch = 1$
		\While{not converged} 
		\State Calculate the prediction ${P_F}\!\left( {{y_i}|{x_i}} \right)$ and loss via forward computation.
		\If{$epoch \% 2 == 1 $} 
		\State Backpropagate the loss to update model parameters $\bf W$.
		\Else
		\State Perform \textsc{AD-Drop} by Eq.~(\ref{eq5})-(\ref{eq8}) to obtain mask matrices ${\bf{M}} = \left[ {{\bf{M}}_1,{\bf{M}}_2, \cdots ,{\bf{M}}_H} \right]$.
		\State Calculate the new prediction ${P_F}\!\left( {{y_i}|{x_i}} \right)$ and new loss by feeding $\bf {M}$ into Eq.~(\ref{eq2}).
		\State Backpropagate the new loss to update model parameters $\bf W$.
		\EndIf
		\State $epoch = epoch + 1$
		\EndWhile
		\State \Return ${\bf {\widetilde {W}}} =\bf W$
	\end{algorithmic}
	\label{alg:cross}
\end{algorithm}

\section{Experiments}\label{experiment}
\subsection{Datasets}
We conduct our main experiments on eight tasks of the GLUE benchmark \cite{wang2018glue}, including SST-2 \cite{socher2013recursive}, MNLI \cite{williams2017broad}, QNLI \cite{rajpurkar2016squad}, QQP \cite{chen2018quora}, CoLA \cite{warstadt2019cola}, STS-B \cite{cer2017semeval}, MRPC \cite{dolan2005automatically}, and RTE \cite{bentivogli2009fifth}. The evaluation metrics are Matthew's Corrcoef (Mcc) \cite{matthews1975comparison} for CoLA, Pearson Corrcoef (Pcc) \cite{benesty2009pearson} for STS-B, and Accuracy (Acc) for the others. To demonstrate that \textsc{AD-Drop}
applies to token-level tasks as well, we conduct experiments on Named Entity Recognition (CoNLL-2003 \cite{sang2003introduction}) and Machine Translation (WMT 2016 \cite{bojar2016findings}) datasets, the results of which are shown in Appendix \ref{apd:tok_exp}. Besides, we also evaluate \textsc{AD-Drop} on two out-of-distribution (OOD) datasets, including HANS \cite{mccoy2019right} and PAWS-X \cite{yang2019paws}. The details of these datasets are introduced in Appendix \ref{apd:exp_dataset}. 

\subsection{Implementation Details}\label{details}
We implement our \textsc{AD-Drop} in Pytorch with the Transformers package \cite{wolf-etal-2020-transformers}.~We train the selected PrLMs on GeForce RTX 3090 GPUs. We tune the learning rate in \{1e-5, 2e-5, 3e-5\} and the batch size in \{16, 32, 64\}. Following \citet{miao2021simple}, we perform early stopping to choose the number of training epochs on GLUE. The two critical hyperparameters $p$ and $q$ are searched within $[0.1, 0.9]$ with step size 0.1. For integrated gradient in Eq.~(\ref{eq3}), we follow \citet{hao2021self} and set $m$ to 20. We apply \textsc{AD-Drop} only in the first layer for the datasets of SST-2, MNLI, QNLI, QQP, and STS-B since the fine-tuning on these datasets is stable and less likely to cause overfitting. For the rest datasets, we apply \textsc{AD-Drop} in all layers. We provide the detailed hyperparameter settings on each dataset in Appendix \ref{apd:exp_set}. Our code is available at \url{https://github.com/TaoYang225/AD-DROP}.

\begin{table*}[t]
	\caption{Overall results of fine-tuned models on the GLUE benchmark. The symbol $\dagger$ denotes results directly taken from the original papers. The best average results are shown in bold.}
	\renewcommand{\arraystretch}{1.0}
	\centering
	\resizebox{1.0\textwidth}{!}{
		\begin{tabular}{lccccccccc}
			\toprule
			Methods & SST-2 & MNLI & QNLI & QQP & CoLA & STS-B & MRPC & RTE & Average\\ 
			\midrule
			\multicolumn{10}{c}{\textit{Development}} \\
			BERT$_{\rm{base}}$ & 92.3 & 84.6 & 91.5 & 91.3 & 60.3 & 89.9 & 85.1 & 70.8 & 83.23 \\
			\ \  +\textit{SCAL}$^\dagger$ \cite{miao2021simple} & 92.8 & 84.1 & 90.9 & 91.4 & 61.7 & - & - & 69.7 & - \\
			\ \  +\textit{SuperT}$^\dagger$ \cite{liang2021super} & 93.4 & 84.5 & 91.3 & 91.3 & 58.8 & 89.8 & 87.5 & 72.5 & 83.64 \\
			\ \  +\textit{R-Drop}$^\dagger$ \cite{wu2021r}  & 93.0 & 85.5 & 92.0 & 91.4 & 62.6 & 89.6 & 87.3 & 71.1 & 84.06 \\
			\ \  +\textsc{AD-Drop}  & 93.9 & 85.1 & 92.3 & 91.8 & 64.6 & 90.4 & 88.5 & 75.1 & \textbf{85.21} \\
			\midrule
			RoBERTa$_{\rm{base}}$ & 95.3 & 87.6 & 92.9 & 91.9 & 64.8 & 90.9 & 90.7 & 79.4 & 86.69 \\
			\ \  +\textit{R-Drop} \cite{wu2021r} & 95.2 & 87.8 & 93.2 & 91.7 & 64.7 & 91.2 & 90.5 & 80.5 & 86.85 \\
			\ \  +\textit{HiddenCut}$^\dagger$ \cite{chen2021hiddencut} & 95.8 & 88.2 & 93.7 & 92.0 & 66.2 & 91.3 & 92.0 & 83.4 & 87.83 \\
			\ \  +\textsc{AD-Drop}  & 95.8 & 88.0 & 93.5 & 92.0 & 66.8 & 91.4 & 92.2 & 84.1 & \textbf{87.98}\\
			\midrule
			\midrule
			\multicolumn{10}{c}{\textit{Test}} \\
			BERT$_{\rm{base}}$ & 93.6 & 84.7 & 90.4 & 89.3 & 52.8 & 85.6 & 81.4 & 68.4 & 80.78 \\
			\ \  +\textsc{AD-Drop}  & 94.3 & 85.2 & 91.6 & 89.4 & 53.3 & 86.6 & 84.1 & 68.7 & \textbf{81.65} \\
			\midrule
			RoBERTa$_{\rm{base}}$ & 94.8 & 87.5 & 92.8 & 89.6 & 58.3 & 88.7 & 86.3 & 75.1 & 84.14 \\
			\ \  +\textsc{AD-Drop}  & 95.9 & 87.6 & 93.4 & 89.5 & 58.5 & 89.3 & 87.9 & 76.0 & \textbf{84.76} \\
			\bottomrule
	\end{tabular}}
	\label{tabel1}
	\vspace{-0.3cm}
\end{table*}

\subsection{Overall Results}
We report the overall results of the fine-tuned models in Table \ref{tabel1}. We first compare \textsc{AD-Drop} with existing regularization methods on the development sets, including the original fine-tuning, SCAL \cite{miao2021simple}, SuperT \cite{liang2021super}, R-Drop  \cite{wu2021r}, and HiddenCut \cite{chen2021hiddencut}. We observe that \textsc{AD-Drop} surpasses the baselines on most of the datasets. Specifically, \textsc{AD-Drop} yields an average improvement of 1.98 and 1.29 points on BERT$_{\rm{base}}$ and RoBERTa$_{\rm{base}}$, respectively. We then discuss the performance of \textsc{AD-Drop} on the test sets. Results in Table \ref{tabel1} show that \textsc{AD-Drop} achieves consistent improvement, boosting the average scores of BERT$_{\rm{base}}$ and RoBERTa$_{\rm{base}}$ by 0.87 and 0.62, respectively. Besides, compared with large datasets, \textsc{AD-Drop} achieves more gains on small datasets, which are more likely to cause overfitting, illustrating that \textsc{AD-Drop} is more suitable for small data scenarios.

\begin{wraptable}{r}{7.7cm}
	\vspace{-0.6cm}
	\small
	\center
	\caption{Results of ablation studies, in which \textit{r/w} means ``replace with'' and \textit{w/o} means ``without''.}
	\begin{tabular}{lcccc}
		\toprule
		Methods & CoLA & STS-B & MRPC & RTE \\ 
		\midrule
		BERT$_{\rm{base}}$ & 60.3 & 89.9 & 85.1 & 70.8 \\
		$\;\,$+\textsc{AD-Drop} (GA) & \textbf{64.6} & 90.4 & \textbf{88.5} & \textbf{75.1} \\
		\quad\textit{r/w} IGA & 63.8 & \textbf{90.7} & \textbf{88.5} & 74.4 \\
		\quad\textit{r/w} AA  & 63.6 & 90.0 & 88.0 & 74.7 \\
		\quad\textit{r/w} RD & 62.1 & 90.2 & 87.8 & 74.7 \\
		\quad\textit{r/w} gold labels & 63.2 & - & 88.0 & 74.4 \\
		\quad\textit{w/o} cross-tuning & 62.1 & 90.4 & 87.3 & 71.5 \\
		\midrule
		RoBERTa$_{\rm{base}}$ & 64.8 & 90.9 & 90.7 & 79.4 \\
		$\;\,$+\textsc{AD-Drop} (GA) & 66.8 & 91.4 & \textbf{92.2} & \textbf{84.1} \\
		\quad\textit{r/w} IGA  & \textbf{68.1} & \textbf{91.6} & 91.4 & 82.7\\
		\quad\textit{r/w} AA  & 66.3 & 91.5 & 91.2 & 82.3 \\
		\quad\textit{r/w} RD & 66.5 & 91.5 & \textbf{92.2} & 82.0 \\
		\quad\textit{r/w} gold labels & 66.4 & - & 91.2 & 82.0 \\
		\quad\textit{w/o} cross-tuning & 67.3 & 91.3 & 90.4 & 80.5 \\
		\bottomrule
	\end{tabular}	
	\label{tabel2}
	\vspace{-0.5cm}
\end{wraptable}

\subsection{Ablation Study}
\label{ablation}
We conduct ablation experiments on four small datasets to investigate the impact of different components. Due to the limited number of submissions imposed by the GLUE server for evaluation, the results here are reported on the development sets.

\paragraph{Attribution methods} \textsc{AD-Drop} can be implemented with different attribution methods to generate the mask matrix in Eq.~(\ref{eq2}), such as integrated gradient attribution (IGA) introduced Eq.~(\ref{eq3}), attention weights for attribution (AA), and randomly generating the discard region (RD) in Eq.~(\ref{eq:discard_region}). We replace the gradient attribution (GA) in Eq.~(\ref{eq:attribution_matrix})-(\ref{eq:discard_region}) with these methods. From Table \ref{tabel2}, we can make three observations. First, \textsc{AD-Drop} with gradient-based attribution methods (GA and IGA) surpasses that with the other methods (AA or RD) on most of the datasets, illustrating that gradient-based methods are better at finding features that are likely to cause overfitting. Second, IGA outperforms GA in some cases. Although IGA provides better theoretical justification than GA for attribution, it requires prohibitively more computational cost than GA (see Section \ref{time} for efficiency analysis), making GA a more desirable choice for \textsc{AD-Drop}. Third, \textsc{AD-Drop} improves the original BERT$_{\rm{base}}$ and RoBERTa$_{\rm{base}}$ with any of the masking strategies, demonstrating the robustness of \textsc{AD-Drop} to overfitting when fine-tuning these models.

\paragraph{Pseudo labels \textit{vs} gold labels} In Section \ref{addrop}, we discuss the motivation of using pseudo labels for attribution in \textsc{AD-Drop}. To verify the reasonability, we conduct an experiment with gold labels for attribution. As the results show in Table \ref{tabel2}, using gold labels for attribution deteriorates the performance, illustrating that \textsc{AD-Drop} with pseudo labels for attribution is preferable.

\begin{wrapfigure}{r}{7cm}
	\vspace{-0.4cm}
	\centering
	\includegraphics[width=0.5\textwidth]{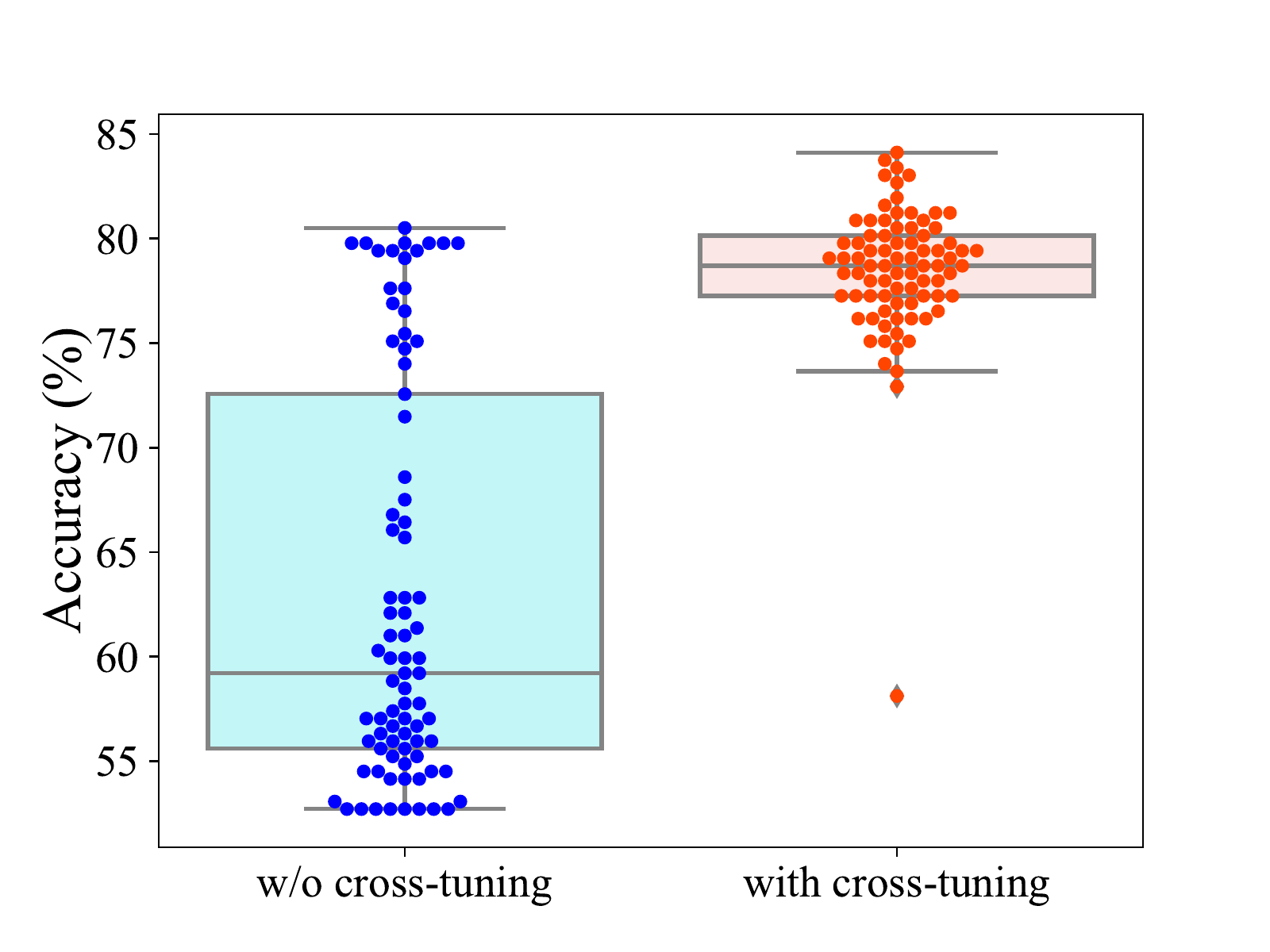}
	\caption{Results of \textsc{AD-Drop} with and without cross-tuning when enumerating $p$ and $q$ in [0.1, 0.9]. RoBERTa is chosen as the base model. Results show that "with cross-tuning" leads to much lower variance and higher performance.}
	\label{fig:wocross}
	\vspace{-0.35cm}
\end{wrapfigure}

\paragraph{Cross-tuning} To verify the effectiveness of the cross-tuning strategy, we ablate it and apply only \textsc{AD-Drop} in all training epochs. As shown in Table \ref{tabel2}, removing cross-tuning causes noticeable performance degradation on most of the datasets.~This can be explained by the intuition that \textsc{AD-Drop} without cross-tuning tends to discard high-attribution positions excessively and make the model difficult to converge normally. To vividly demonstrate the effect of \textsc{AD-Drop} with or without cross-tuning, we visualize the distributions of the performance on the RTE\footnote{Results on the other datasets are shown in Appendix \ref{apd:results_cross}.} development set when enumerating the parameters $p$ and $q$ in the range of [0.1, 0.9]. The results are plotted in Figure \ref{fig:wocross}, where each blue/orange point denotes the accuracy with a pair of $p$ and $q$ values. We observe from the figure that \textsc{AD-Drop} without cross-tuning cannot be trained properly under some parameter settings. However, it works well for most parameter settings when cross-tuning is applied, demonstrating that cross-tuning is vital for improving training stability. 

\section{Analysis} \label{analysis}
In this section, we further conduct several experiments for more thorough analysis. 

\begin{wraptable}{r}{8.5cm}
\vspace{-0.55cm}
	\small
	\center
	\caption{Results of repeated experiments. Each score is the average of five runs with a standard deviation.}
	\begin{tabular}{lcccc}
		\toprule
		Methods & CoLA & STS-B & MRPC & RTE \\ 
		\midrule
		BERT$_{\rm{base}}$ & 61.8$_{\pm 1.9}$ & 89.4$_{\pm\bf 0.5}$ & 85.2$_{\pm1.3}$ & 71.2$_{\pm1.2}$ \\
		$\;\;$+\textsc{AD-Drop} & \textbf{63.4}$_{\pm \bf 0.4}$ & \textbf{90.1}$_{\pm \bf 0.5}$ & \textbf{87.4}$_{\pm \bf 0.9}$ & \textbf{73.9}$_{\pm \bf 1.1}$ \\
		\midrule
		RoBERTa$_{\rm{base}}$ & 64.3$_{\pm \bf 0.9}$ & 91.0$_{\pm 0.2}$ & 89.8$_{\pm 0.8}$ & 79.1$_{\pm 1.7}$ \\
		$\;\;$+\textsc{AD-Drop} & \textbf{66.4}$_{\pm \bf 0.9}$ & \textbf{91.2}$_{\pm \bf 0.1}$ &  \textbf{91.3}$_{\pm \bf 0.7}$ & \textbf{82.5}$_{\pm \bf 0.9}$ \\
		\bottomrule
	\end{tabular}	
	\label{tabel:repeat}
	\vspace{-0.3cm}
\end{wraptable}

\subsection{Repeated Experiments} \label{repeat}
To reduce the influence of randomness, we conduct repeated experiments on four small datasets (i.e., CoLA, STS-B, MRPC and RTE). We repeat the training of each model with five random seeds and report the average score and standard deviation on the development sets. From Table \ref{tabel:repeat}, we observe that \textsc{AD-Drop} outperforms the original fine-tuning on all the datasets. In addition, \textsc{AD-Drop} results in lower standard deviations on most of the datasets, showing that \textsc{AD-Drop} is more robust in fine-tuning PrLMs than the original approach.

\subsection{Effect of Data Size}
\begin{wrapfigure}{r}{5.8cm}
    \vspace{-0.0cm}
	\centering
	\includegraphics[width=0.4\textwidth]{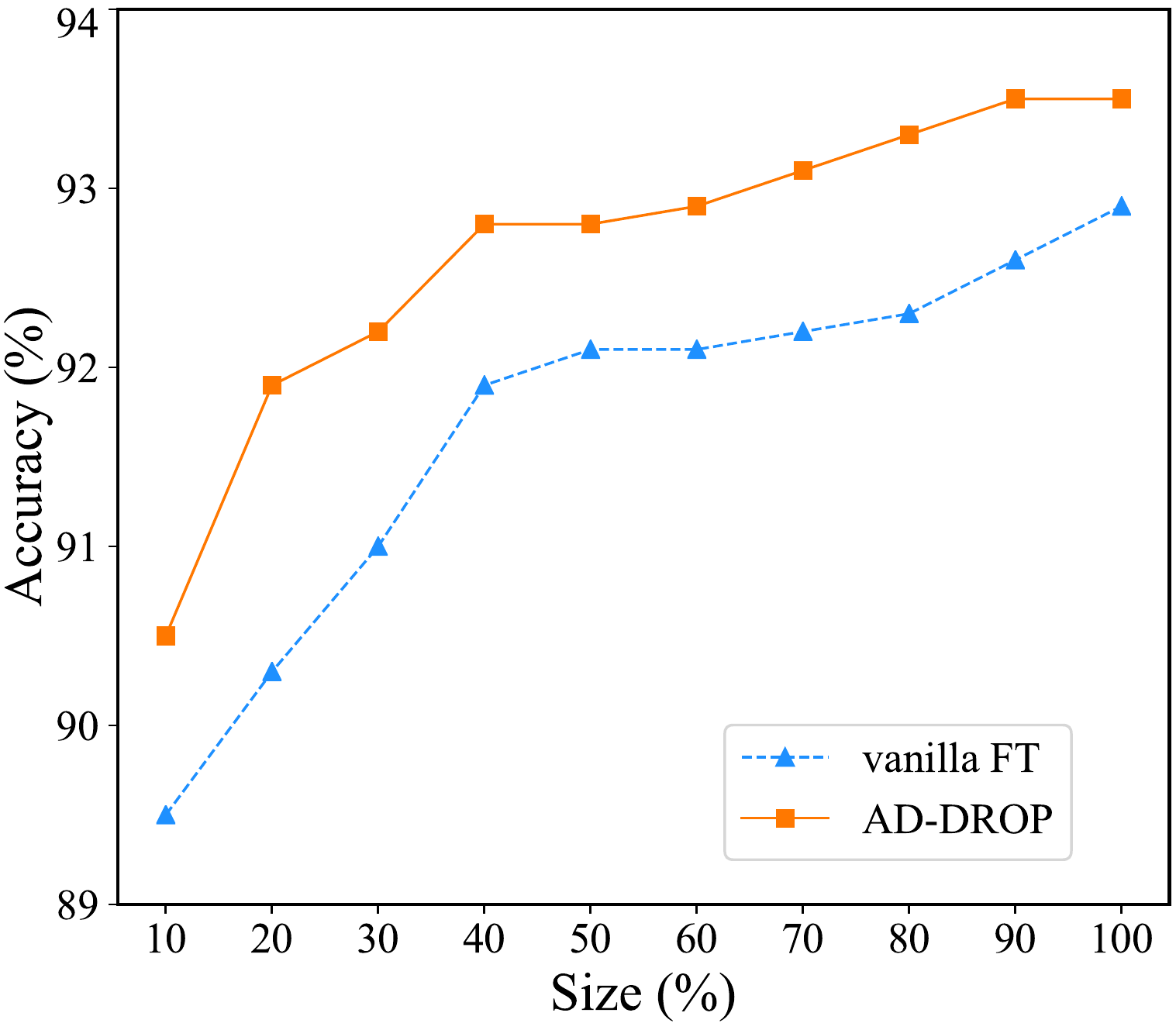}
	\caption{Results of \textsc{AD-Drop} and FT as the number of training samples changes.}
	\label{fig:size}
	\vspace{-1cm}
\end{wrapfigure}

To study the impact of data size, we compare \textsc{AD-Drop} with the original fine-tuning (FT) approach on QNLI and QQP,\footnote{Results on QQP are shown in Appendix \ref{apd:qqp}.} two relatively large datasets, and report their performance when the number of training samples changes. RoBERTa is chosen as the base model. Figure \ref{fig:size} shows that \textsc{AD-Drop}  outperforms FT consistently on QNLI. Moreover, \textsc{AD-Drop} improves the efficiency of data use as training \textsc{AD-Drop} with 60\% training samples produces comparable performance to FT with full data.

\subsection{Hyperparameter Sensitivity}
\begin{figure*}[htbp]
	\centering
	\subfloat[BERT on CoLA]{\label{fig_vis_a}
		\includegraphics[width=0.228\textwidth]{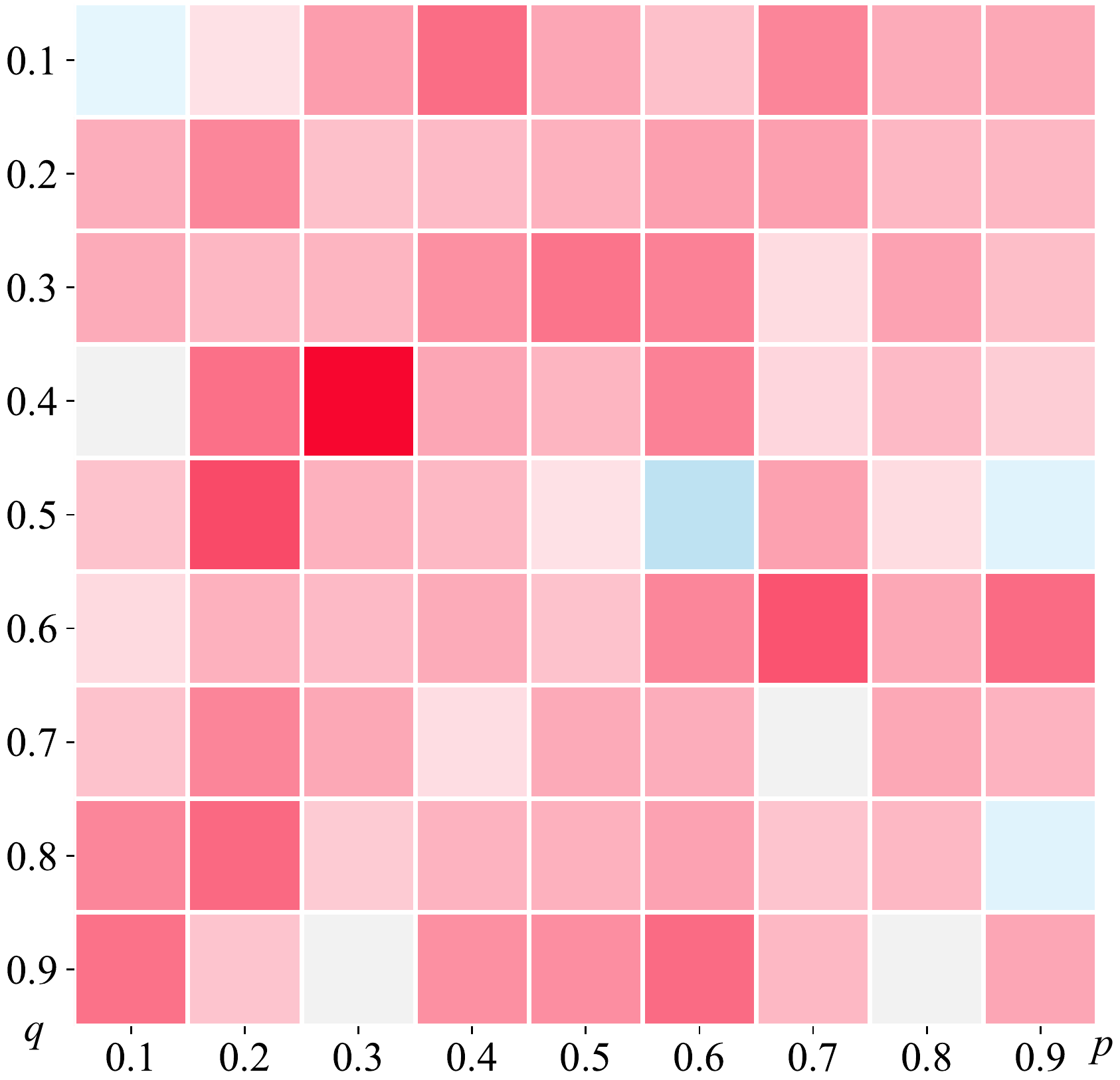}}\hspace{-0.1mm}
	\subfloat[RoBERTa on CoLA]{\label{fig_vis_b}
		\includegraphics[width=0.228\textwidth]{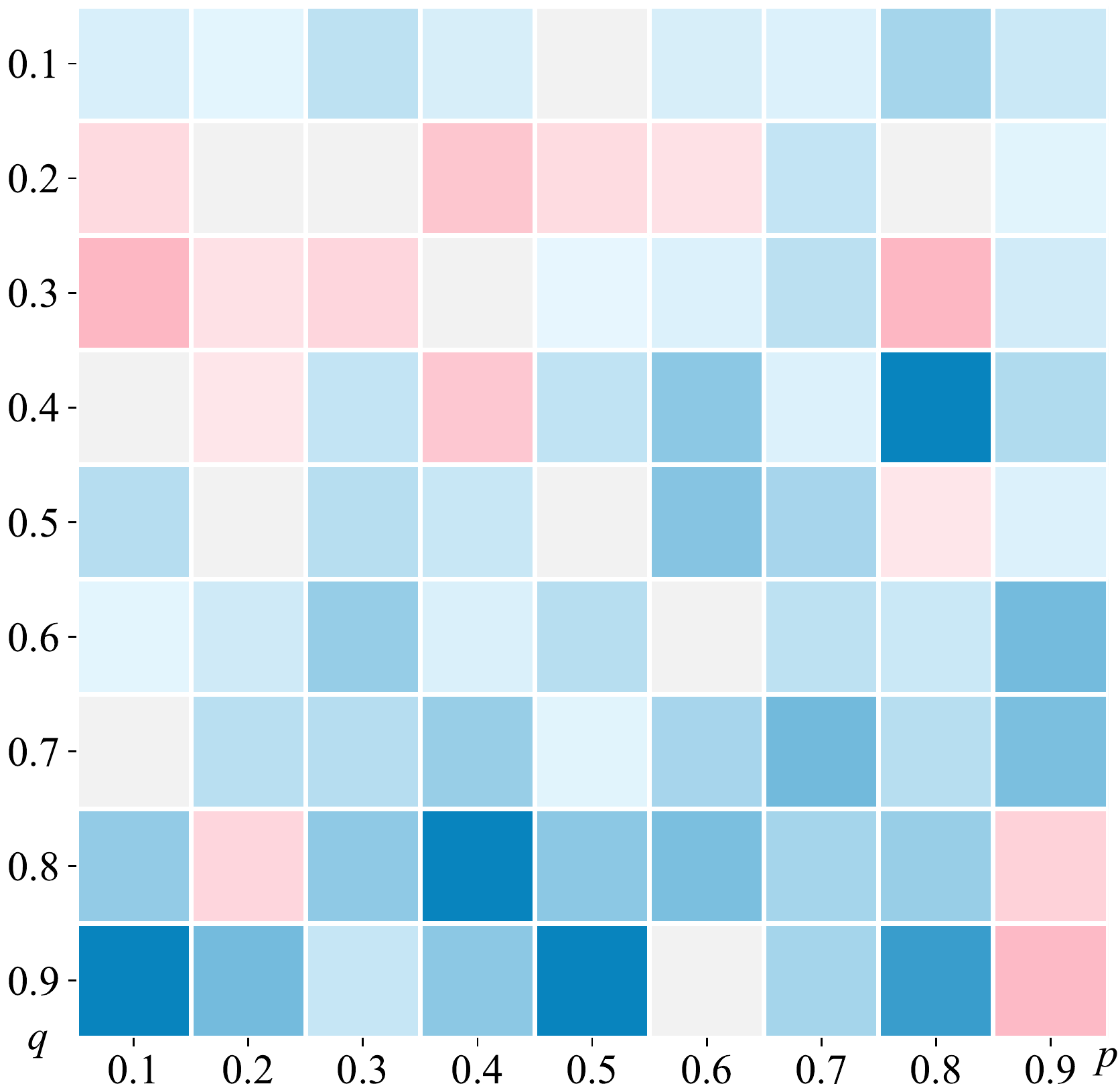}}\hspace{-0.1mm}
	\subfloat[BERT on RTE]{\label{fig_vis_c}
		\includegraphics[width=0.228\textwidth]{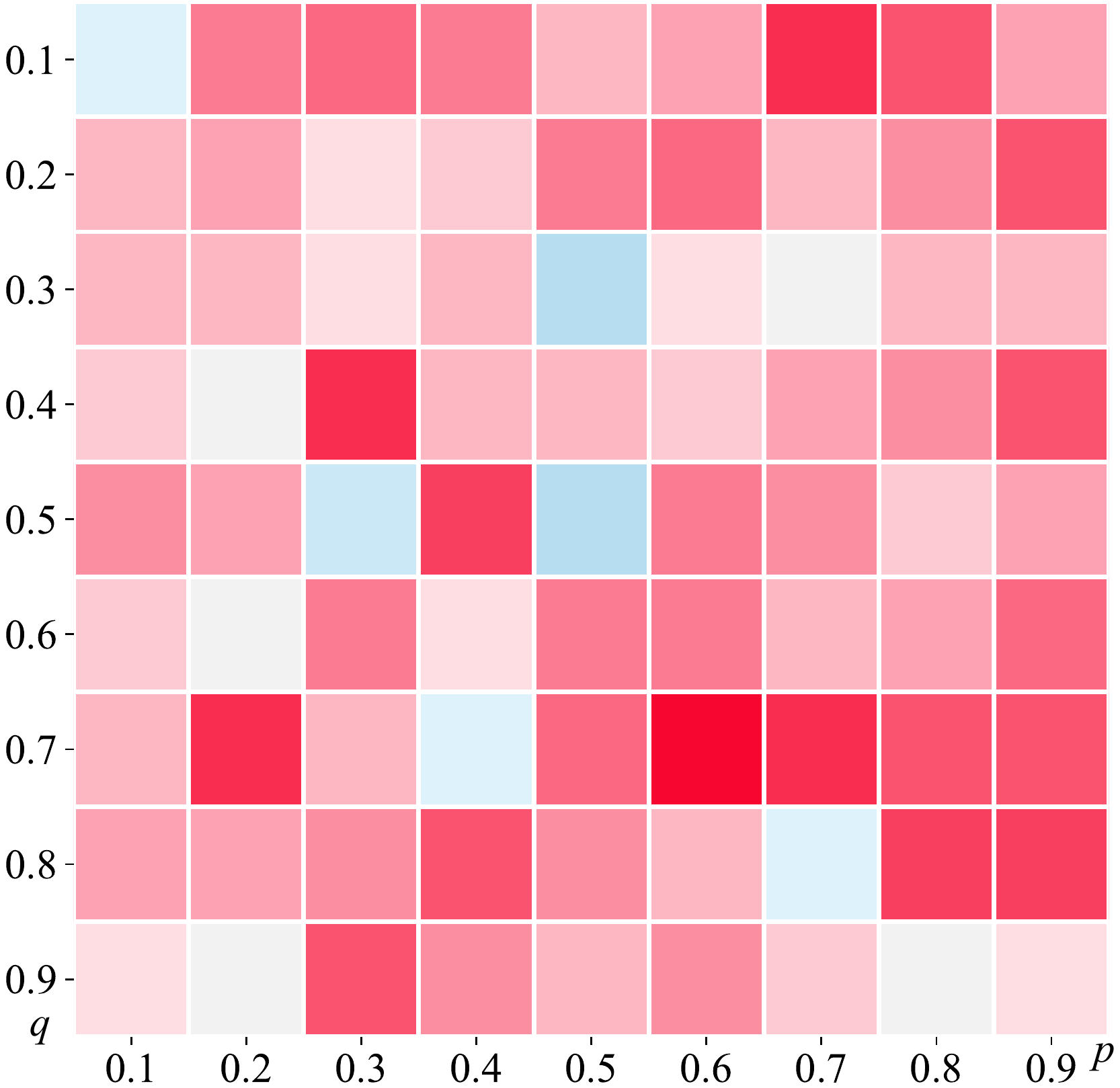}}\hspace{-0.1mm}
	\subfloat[RoBERTa on RTE]{\label{fig_vis_d}
		\includegraphics[width=0.268\textwidth]{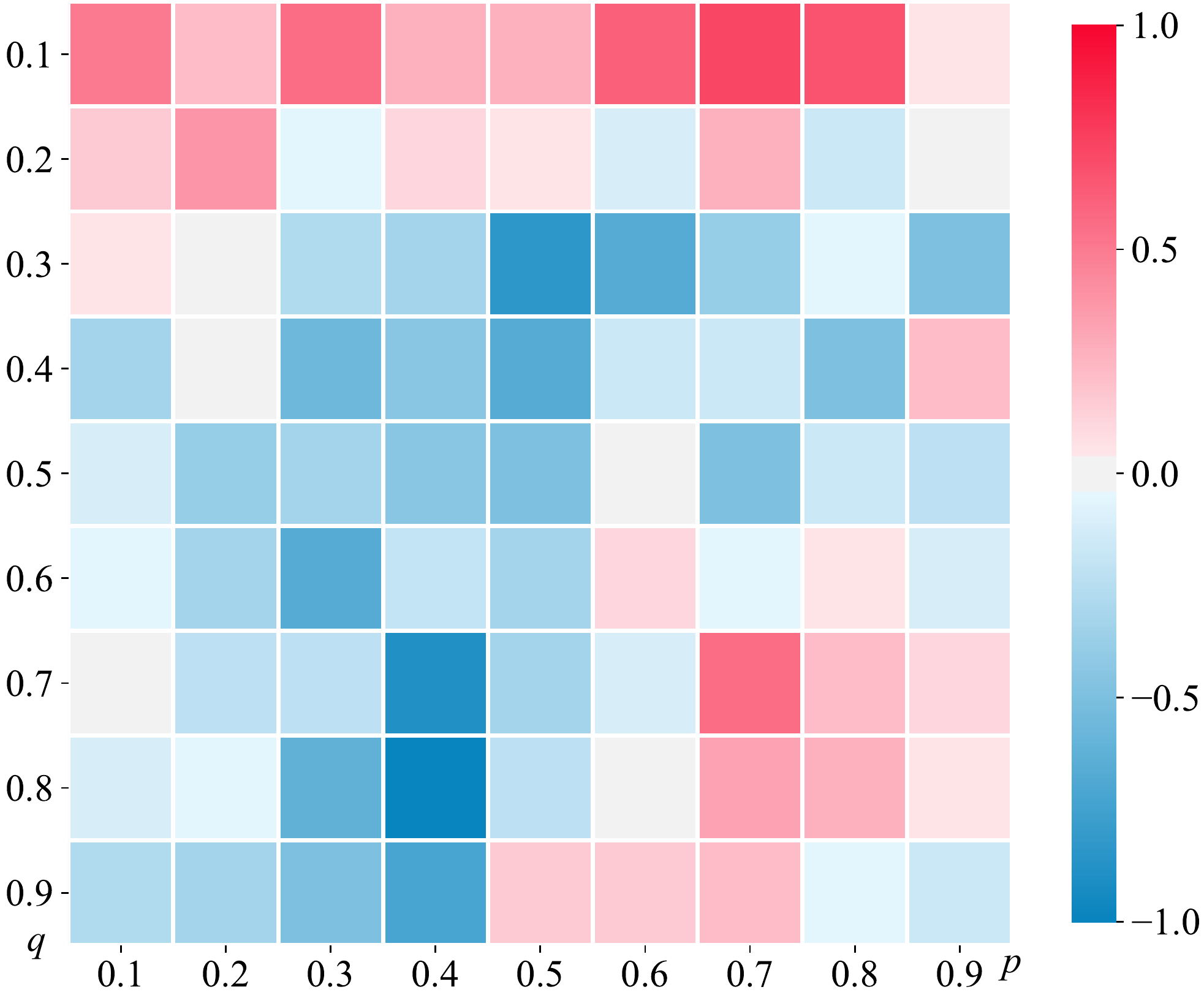}}
		\centering
	\caption{Results of sensitivity study on CoLA and RTE. Rows correspond to $p$ and columns refer to $q$. Blue blocks indicate the results of \textsc{AD-Drop} below the baseline (FT), and red blocks mean the results of \textsc{AD-Drop} above the baseline. Darker colors mean greater gaps with the baseline.}
	\label{fig:valid}
\end{figure*}

\textsc{AD-Drop} involves two hyperparameters $p$ and $q$ to control the number of discarded attention positions. To investigate the sensitivity of \textsc{AD-Drop} to them, we show the results of different $p$ and $q$ combinations on CoLA and RTE in Figure \ref{fig:valid}, in which we apply \texttt{MaxAbsScaler}\footnote{https://scikit-learn.org/stable/modules/generated/sklearn.preprocessing.MaxAbsScaler.html} to project the difference between the results of \textsc{AD-Drop} and FT into the interval of $[-1.0, 1.0]$. We observe that BERT with \textsc{AD-Drop} is not hyperparameter-sensitive as it outperforms the baseline under most settings. In contrast, RoBERTa with \textsc{AD-Drop} is more sensitive and requires a careful search for optimal hyperparameter settings. The possible reason is that RoBERTa is pre-trained with more data and more effective tasks than BERT, making it less prone to overfitting than BERT.

\begin{wraptable}{r}{6.5cm}
	\centering
	\vspace{-0.4cm}
	\caption{Testing \textsc{AD-Drop} on a larger model.}
		\begin{tabular}{lcc}
			\toprule
			Methods &  MRPC & RTE \\
			\midrule
			RoBERTa$_{\rm{large}}$ &  90.83$_{\pm 0.75}$ & 85.99$_{\pm 0.86}$  \\
			$\;\;$+\textsc{AD-Drop} &  \textbf{91.62}$_{\pm \bf 0.53}$ & \textbf{88.01}$_{\pm \bf 0.48}$  \\
			\bottomrule
		\end{tabular}
	\label{tab:larger_model}
\end{wraptable}

\subsection{Larger Model Size}\label{large_model}
To verify the scalability of \textsc{AD-Drop} for a larger model size, we evaluate \textsc{AD-Drop} with RoBERTa$_{\rm{large}}$ on the RTE and MRPC datasets. Table \ref{tab:larger_model} shows the average scores and standard deviations of five random seeds. There are two main observations. First, \textsc{AD-Drop} achieves consistent improvements over the larger RoBERTa model, illustrating that \textsc{AD-Drop} is scalable to large models. Second, compared with RoBERTa$_{\rm{base}}$ on RTE in Table \ref{tabel:repeat}, the larger model significantly reduces the deviation (from 1.7 to 0.86), suggesting that a larger model size indeed helps to improve the stability. \textsc{AD-Drop} further improves the performance and reduces the deviation.

\subsection{Few-shot Scenario }\label{few_shot}
In this subsection, we test the performance of \textsc{AD-Drop} under few-shot scenarios. Specifically, we carry out 16-, 64-, and 256-shot experiments on SST-2 and CoLA with RoBERTa$_{\rm{base}}$ as the base model and the baseline. We report the average scores and standard deviations of five random seeds in Table \ref{tabel:fewshot}. We observe that RoBERTa with \textsc{AD-Drop} consistently outperforms the original fine-tuning approach. Besides, \textsc{AD-Drop} tends to bring more benefits when fewer samples are available.

\begin{table*}[htbp]
	\small
	\center
	\caption{Testing \textsc{AD-Drop} in few-shot settings. RoBERTa with \textsc{AD-Drop} achieves higher performance and lower deviations than that with the original fine-tuning approach.}
	\resizebox{1.0\textwidth}{!}{
	\begin{tabular}{l|ccc|ccc}
		\toprule
		\multirow{2}{*}{Methods} & \multicolumn{3}{c|}{SST-2} & \multicolumn{3}{c}{CoLA}\\ 
		& 16-shot & 64-shot & 256-shot & 16-shot & 64-shot & 256-shot \\
		\midrule
		RoBERTa$_{\rm{base}}$ & 74.50$_{\pm 3.03}$ & 89.06$_{\pm 0.83}$ & 91.44$_{\pm 0.17}$ & 23.18$_{\pm 6.38}$ & 39.70$_{\pm 4.68}$ & 51.11$_{\pm 1.64}$ \\
		$\;\;$+\textsc{AD-Drop} & \textbf{80.16}$_{\pm \bf 1.51}$ & \textbf{91.61}$_{\pm \bf 0.52}$ & \textbf{92.61}$_{\pm \bf 0.13}$ & \textbf{26.70}$_{\pm \bf 4.96}$ & \textbf{46.41}$_{\pm \bf 1.98}$ & \textbf{52.47}$_{\pm \bf 1.16}$ \\
		\bottomrule
	\end{tabular}}	
	\label{tabel:fewshot}
\end{table*}

\begin{wraptable}{r}{6.5cm}
	\centering
	\vspace{-0.4cm}
	\caption{Testing \textsc{AD-Drop} on OOD datasets.}
		\begin{tabular}{lcc}
			\toprule
			Methods &  HANS & PAWS-X \\
			\midrule
			RoBERTa$_{\rm{base}}$ &  69.83 & 47.90  \\
			$\;\;$+\textsc{AD-Drop} &  \textbf{70.49} & \textbf{51.25}  \\
			\bottomrule
		\end{tabular}
	\label{tab:ood}
\end{wraptable}

\subsection{Out-of-Distribution Generalization} \label{ood_test}
To further demonstrate \textsc{AD-Drop} is beneficial to reducing overfitting, we test \textsc{AD-Drop} with RoBERTa$_{\rm{base}}$ on two out-of-distribution (OOD) datasets, i.e., HANS and PAWS-X. For HANS, we use the checkpoints trained on MNLI and test their performance on the validation set (the test set is not supplied). For PAWS-X, we use the checkpoints trained on QQP and examine its performance on the test set. The evaluation metric is accuracy. From Table \ref{tab:ood}, we can see that RoBERTa with \textsc{AD-Drop} achieves better generalization, where \textsc{AD-Drop} boosts the performance by 0.66 on HANS and 3.35 on PAWS-X, illustrating that the model trained with \textsc{AD-Drop} generalizes better to OOD data.

\subsection{Computational Efficiency}\label{time}
To analyze the computational efficiency, we quantitatively study the computational cost of \textsc{AD-Drop} with different dropping strategies (GA, IGA, AA, and RD) relative to the original fine-tuning on CoLA, STS-B, MRPC, and RTE. BERT is chosen as the base model for this experiment. As shown in Table \ref{tabel:time}, although IGA achieves more favorable performance on one of the datasets, it requires higher computational costs than its counterparts, especially when applied in all the layers. In contrast, \textsc{AD-Drop} with GA is more competitive in both performance and computational cost. 

\begin{table*}[htbp]
	\small
	\center
	\caption{Results of performance and computational cost of \textsc{AD-Drop} with different masking strategies (GA, IGA, AA, and RD) relative to the original fine-tuning. The symbol $\ddagger$ means \textsc{AD-Drop} is only applied in the first layer. BERT is chosen as the base model.}
	\resizebox{.9\textwidth}{!}{
	\begin{tabular}{c|cc|cc|cc|cc}
		\toprule
		\multirow{2}{*}{Methods} & \multicolumn{2}{c|}{CoLA} & \multicolumn{2}{c|}{STS-B$^\ddagger$} & \multicolumn{2}{c|}{MRPC} & \multicolumn{2}{c}{RTE} \\ 
		& Mcc & Time & Pcc & Time & Acc & Time & Acc & Time \\
		\midrule
		RD & +1.8 & $\times$1.42 & +0.3 & $\times$1.38 & +2.7 & $\times$1.31 & +3.9 & $\times$1.42 \\
		AA & +3.3 & $\times$1.42 & +0.1 & $\times$1.48 & +2.9 & $\times$1.94 & +3.9 & $\times$1.58 \\
		GA & +4.3 & $\times$3.58 & +0.5 & $\times$1.95 & +3.4 & $\times$4.13 & +4.3 & $\times$4.50 \\
		IGA & +3.5 & $\times$99.61 & +0.8 & $\times$15.00 & +3.4 & $\times$110.12 & +3.6 & $\times$125.67 \\
		\bottomrule
	\end{tabular}}	
	\label{tabel:time}
\end{table*}

\section{Related Work}

\paragraph{Dropout}
Dropout is a widely used regularizer to mitigate overfitting when training deep neural networks. Vanilla dropout \cite{srivastava2014dropout} randomly selects neurons with a predefined probability and sets their values to zeros during training. By doing so, the neurons cannot co-adapt and the trained networks can lead to better generalization. In recent years, many variants of dropout have emerged. One body of research aims to adopt different strategies to drop units in neural networks. For example, DropConnect \cite{wan2013regularization} randomly selects connections between neurons to discard. DropBlock \cite{ghiasi2018dropblock} defines a structured dropout that randomly drops the units in a specific contiguous region of a feature map. AutoDropout \cite{pham2021autodropout} aims to improve the dropout pattern of DropBlock by introducing an automatic method to design dropout structures. HiddenCut \cite{chen2021hiddencut} drops contiguous spans within the hidden space, in which the attention weights are utilized to select the dropped spans strategically.

Another body of research devotes to addressing the inconsistency between training and inference when dropout is applied. For instance, mixout \cite{lee2019mixout} randomly replaces selected parameters with original pre-trained weights rather than setting them to zeros. CHILD-TUNING \cite{xu2021raise} selects a child network and masks out the gradients of the non-child network during the backward step, only updating weights in the child network. R-Drop \cite{wu2021r} performs dropout twice in the forward steps to produce two sub-models and then applies KL-divergence for their output distributions, forcing the two sub-models to be consistent with each other. However, most of these methods follow the random sampling strategy of dropout and pay little attention to the different importance of self-attention positions in PrLMs.

\paragraph{Attribution}
Numerous studies have been undertaken to interpret the behaviors of deep neural networks (DNNs). As a theory for interpretability, attribution aims to evaluate the impact of input features on predictions \cite{zhang2021survey}. Generally, attribution methods can be divided into perturbation-based \cite{li2016understanding, feng2018pathologies, Schulz2020Restricting}, gradient-related \cite{baehrens2010explain, binder2016layer, sundararajan2017axiomatic}, and attention-based \cite{rocktaschel2015reasoning,serrano2019attention, wiegreffe2019attention} methods. We focus on reviewing the gradient-related methods as they are more relevant to our work. Specifically, earlier works \cite{baehrens2010explain, simonyan2013deep, springenberg2014striving} try to explain model decisions via gradients since gradients indicate the direction and rate that changes the loss the fastest. However, \citet{sundararajan2017axiomatic} point out that gradient attribution violates the sensitivity axiom in some cases that the gradients will be zero for the function in saturated areas, and propose integrated gradient as a theoretically more sound attribution method.

Other efforts have been devoted to revealing the behavior patterns of PrLMs. \citet{kovaleva2019revealing} and \citet{clark2019does} use attention weights for attribution to investigate what specific knowledge BERT \cite{devlin2018bert} learns. \citet{jain2019attention} and \citet{ brunner2020identifiability} investigate the identifiability of attention weights and conclude that attention weights are not a faithful explanation for model predictions. \citet{hao2021self} apply integrated gradient \cite{sundararajan2017axiomatic} as a saliency measure for self-attention attribution in BERT, and use the attribution result to interpret information interactions inside Transformers. Similarly, \citet{lu2021influence} develop influence patterns based on integrated gradient to explain information flow in BERT. Unlike these works, we aim to examine the effect of dropout on self-attention through self-attention attribution and to reduce overfitting when fine-tuning PrLMs.

\section{Conclusion}
We propose a novel dropout regularizer, \textsc{AD-Drop}, to mitigate overfitting when fine-tuning PrLMs on downstream tasks. Unlike previous dropout-based methods that generally adopt the random sampling strategy to discard units, \textsc{AD-Drop} draws inspiration from self-attention attribution which reveals that attention positions are not equally important in reducing overfitting and that dropping inappropriate positions may exacerbate the problem. Therefore, \textsc{AD-Drop} focuses on discarding high-attribution attention positions to prevent the model from relying heavily on these positions to make predictions. Besides, we propose a cross-tuning strategy that performs the original fine-tuning and our \textsc{AD-Drop} alternately to stabilize the fine-tuning process. Extensive experiments and analysis on the GLUE benchmark demonstrate the effectiveness of \textsc{AD-Drop}. Although originally proposed and evaluated based on self-attention attribution, \textsc{AD-Drop} can be potentially extended to other neural network units as vanilla dropout, which deserves further research efforts.

\section*{Acknowledgments}
We appreciate the anonymous reviewers for their valuable comments. We thank Feifan Yang, Zihong Liang, and Hong Ding for their early discussions. This work was supported by the National Natural Science Foundation of China (No. 62176270), the Program for Guangdong Introducing Innovative and Entrepreneurial Teams (No. 2017ZT07X355), and the Foundation of Key Laboratory of Machine Intelligence and Advanced Computing of the Ministry of Education.

\bibliographystyle{unsrtnat}
\bibliography{ref}

\clearpage

\section*{Checklist}

\begin{enumerate}

\item For all authors...
\begin{enumerate}
  \item Do the main claims made in the abstract and introduction accurately reflect the paper's contributions and scope?
    \answerYes{Our contributions and scope are accurately reflected in the abstract and introduction.}
  \item Did you describe the limitations of your work?
    \answerYes{We discuss the potential limitations of our work in Appendix \ref{apd:limit}.}
  \item Did you discuss any potential negative societal impacts of your work?
    \answerNo{Our work is a general training method and will not bring any negative societal impact.}
  \item Have you read the ethics review guidelines and ensured that your paper conforms to them?
    \answerYes{We have read the guidelines and ensured our paper conforms to them.}
\end{enumerate}

\item If you are including theoretical results...
\begin{enumerate}
  \item Did you state the full set of assumptions of all theoretical results?
    \answerNA{}
        \item Did you include complete proofs of all theoretical results?
    \answerNA{}
\end{enumerate}

\item If you ran experiments...
\begin{enumerate}
  \item Did you include the code, data, and instructions needed to reproduce the main experimental results (either in the supplemental material or as a URL)?
    \answerYes{We have submitted our code and data in the supplemental material.}
  \item Did you specify all the training details (e.g., data splits, hyperparameters, how they were chosen)?
    \answerYes{We describe our training details in Section \ref{details} and Appendix \ref{apd:exp}.}
        \item Did you report error bars (e.g., with respect to the random seed after running experiments multiple times)?
    \answerYes{We conduct repeated experiments in Section \ref{repeat}.}
        \item Did you include the total amount of compute and the type of resources used (e.g., type of GPUs, internal cluster, or cloud provider)?
    \answerYes{We describe the type of GPUs used in this work in Section \ref{details}.}
\end{enumerate}

\item If you are using existing assets (e.g., code, data, models) or curating/releasing new assets...
\begin{enumerate}
  \item If your work uses existing assets, did you cite the creators?
    \answerYes{We provide appropriate citations for the used data and tools.}
  \item Did you mention the license of the assets?
    \answerNo{All used datasets are publicly available.}
  \item Did you include any new assets either in the supplemental material or as a URL?
\answerNA{}
  \item Did you discuss whether and how consent was obtained from people whose data you're using/curating?
    \answerNA{}
  \item Did you discuss whether the data you are using/curating contains personally identifiable information or offensive content?
    \answerNA{}
\end{enumerate}

\item If you used crowdsourcing or conducted research with human subjects...
\begin{enumerate}
  \item Did you include the full text of instructions given to participants and screenshots, if applicable?
    \answerNA{}
  \item Did you describe any potential participant risks, with links to Institutional Review Board (IRB) approvals, if applicable?
    \answerNA{}
  \item Did you include the estimated hourly wage paid to participants and the total amount spent on participant compensation?
    \answerNA{}
\end{enumerate}

\end{enumerate}


\clearpage

\appendix
\section{Appendix: \textsc{AD-Drop} for Token-Level Tasks}\label{apd:tok}
\subsection{Attribution Matrix}\label{apd:tok_der}
Note that \textsc{AD-Drop} is naturally suitable for classification tasks since we can obtain one single attribution matrix with respect to the only logit output for each attention map. For token-level tasks (e.g., NER and text generation), as we have several logit outputs to produce the corresponding attribution matrices for each attention map, applying \textsc{AD-Drop} has the challenge of how to fuse these attribution matrices. We provide a simple alternative to calculate the attribution matrix in Eq.~(\ref{eq:attribution_matrix}) as:
\begin{equation} \label{eq:attribution_matrix_apd}
{{\widetilde{\bf{B}}}_h = -{\frac{{\partial \mathcal{L}}}{{\partial {\bf{A}}_h}}}},
\end{equation}
where $\mathcal{L}$ is the pseudo loss in terms of the pseudo labels. Given a sequence $x$ with $n$ input tokens, we represent each pseudo label as a one-hot vector of $C$ elements and compute $\mathcal{L}$ as:
\begin{equation} \label{eq:lossfunc_apd}
{\mathcal{L}={\sum_{i=1}^{n}{\mathcal{L}_i} = -{\sum_{i=1}^{n}}{\sum_{c=1}^{C}{y_{i,c}}{\rm{log}}{P_F}\!\left( {{c}|{x,i}} \right) } = -{\sum_{i=1}^{n}}y_{i,\tilde {c}}{\rm{log}}{P_F}\!\left( {\tilde {c}|{x,i}} \right)}},
\end{equation}
where $y_{i,c}$ is the $c$-th element in the one-hot vector for token $i$, ${P_F}\!\left( {{c}|{x,i}} \right)$ is the softmax output of class $c$ for token $i$, and $\tilde{c}$ is the pseudo label. Then, Eq. (\ref{eq:attribution_matrix_apd}) can be updated as:
\begin{equation} \label{eq:derivation_end_apd}
{\widetilde{\bf{B}}}_h = -
{{\frac{{\partial \mathcal{L}}}{{\partial {\bf{A}}_h}}} = -{\sum_{i=1}^{n}}{\frac{{\partial \mathcal{L}_i}}{{\partial F_{i,{\tilde {{c}}}} \left( {{\bf{A}}} \right)}}} \cdot {\frac{{\partial F_{i,{\tilde {{c}}}} \left( {{\bf{A}}} \right)}}{{\partial {\bf{A}}_h}}} = {\sum_{i=1}^{n}}{\left(y_{i,\tilde {c}}- {P_F}\!\left( {\tilde {c}|{x,i}} \right)\right) {\bf{B}}_{i,h}}}.
\end{equation}

Therefore, we can use Eq. (\ref{eq:derivation_end_apd}) to compute a single attribution matrix for each attention map when applying \textsc{AD-Drop} in token-level tasks. Besides, as regression tasks (e.g., STS-B) cannot infer pseudo labels, we directly use the actual loss instead.

\subsection{Token-Level Experiments}\label{apd:tok_exp}
We conduct additional experiments of \textsc{AD-Drop} on NER (CoNLL-2003) and Machine Translation (WMT 2016) tasks.\footnote{We follow the official colab implementation (\url{https://huggingface.co/transformers/v4.7.0/notebooks.html}) for the two tasks.} The results on the test sets are reported in Table \ref{tab:ner} and Table \ref{tab:mt}. Moreover, to verify that \textsc{AD-Drop} can be adapted to other pre-trained models, for CoNLL-2003 NER, we choose ELECTRA as the base model. For WMT 2016, OPUS-MT is chosen. The results show that \textsc{AD-Drop} consistently improves the baselines on both NER and Machine Translation tasks.

\begin{wraptable}{l}{6.5cm}
	\centering
	\caption{Test results of \textsc{AD-Drop} on the CoNLL-2003 NER dataset.}
		\begin{tabular}{lcc}
			\toprule
			Methods &  Accuracy & F1 \\
			\midrule
			ELECTRA$_{\rm{base}}$ &  97.83 & 91.23  \\
			$\;\;$+\textsc{AD-Drop} &  \textbf{97.95} & \textbf{91.77}  \\
			\bottomrule
		\end{tabular}
	\label{tab:ner}
\end{wraptable}

\begin{wraptable}{r}{6.5cm}
	\centering
	\vspace{-4.4cm}
	\caption{Test results of \textsc{AD-Drop} on Translation datasets. The evaluation metric is BLEU.}
	\begin{tabular}{lcc}
		\toprule
		Methods &  EN-RO & TR-EN \\
		\midrule
		OPUS-MT &  26.11 & 23.88  \\
		$\;\;$+\textsc{AD-Drop} &  \textbf{26.43} & \textbf{23.96}  \\
		\bottomrule
	\end{tabular}
	\label{tab:mt}
\end{wraptable}

\section{Appendix: More Prior Experiments}\label{apd:probe}

\begin{figure*}[htbp]
	\centering
	\includegraphics[width=1.0\textwidth]{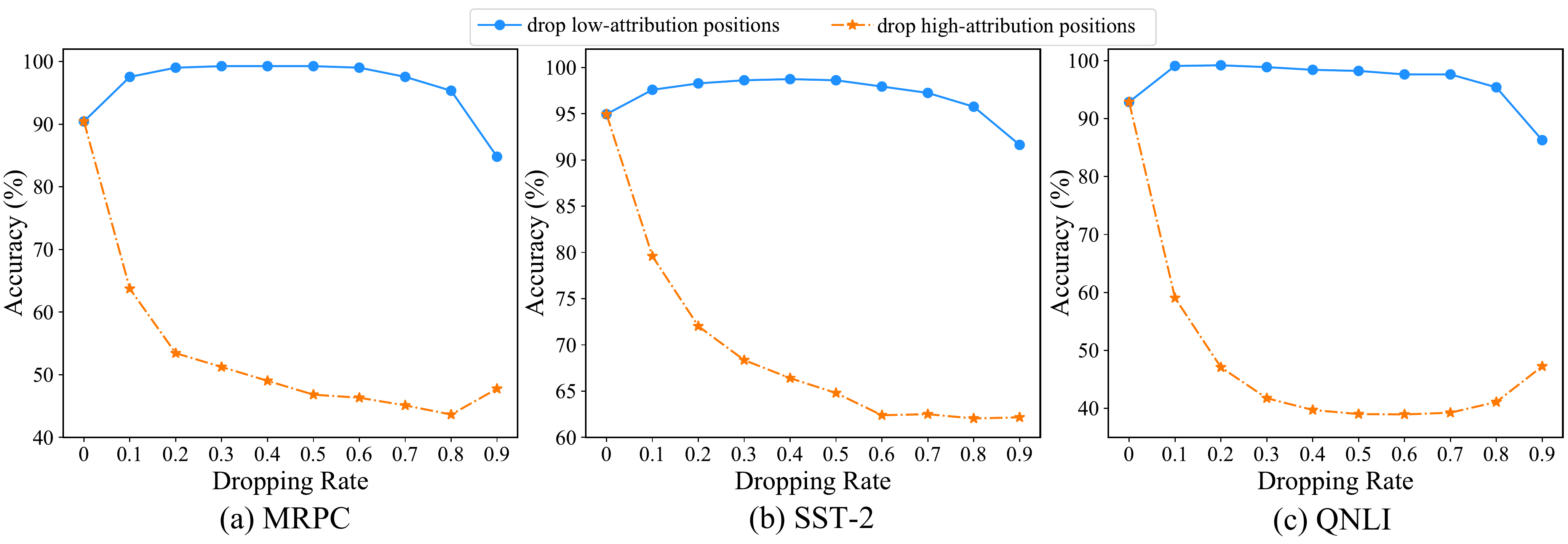}
	
	\caption{Performance of fine-tuned RoBERTa on development sets, where two dropping strategies (i.e., drop low-/high-attribution positions) are applied. Gold labels are used for the attribution.}
	\label{fig:appendix_prior}
\end{figure*}

Our observations in Figure \ref{fig:prior_results}(a) show that dropping low-attribution positions makes the model fit the training data rapidly, while dropping high-attribution positions reduces the fitting speed. To further probe the effect of dropping low- or high-attention positions, we fine-tune a RoBERTa on the training set and evaluate its performance on the development set by applying the two dropping strategies. The results on MRPC, SST-2, and QNLI are plotted in Figure \ref{fig:appendix_prior}. Similar phenomena can be observed that the model rapidly fits the data while dropping only a small proportion of low-attribution positions. As the dropping rate increases, the accuracy remains stable until discarding too many low-attribution positions. When dropping high-attribution positions, we observe an opposite trend that the performance deteriorates sharply. These results further 
confirm the observations in Section \ref{prior} that attention positions should not be treated equally important in dropout. 

\begin{figure*}[htbp]
	\includegraphics[width=1.0\textwidth]{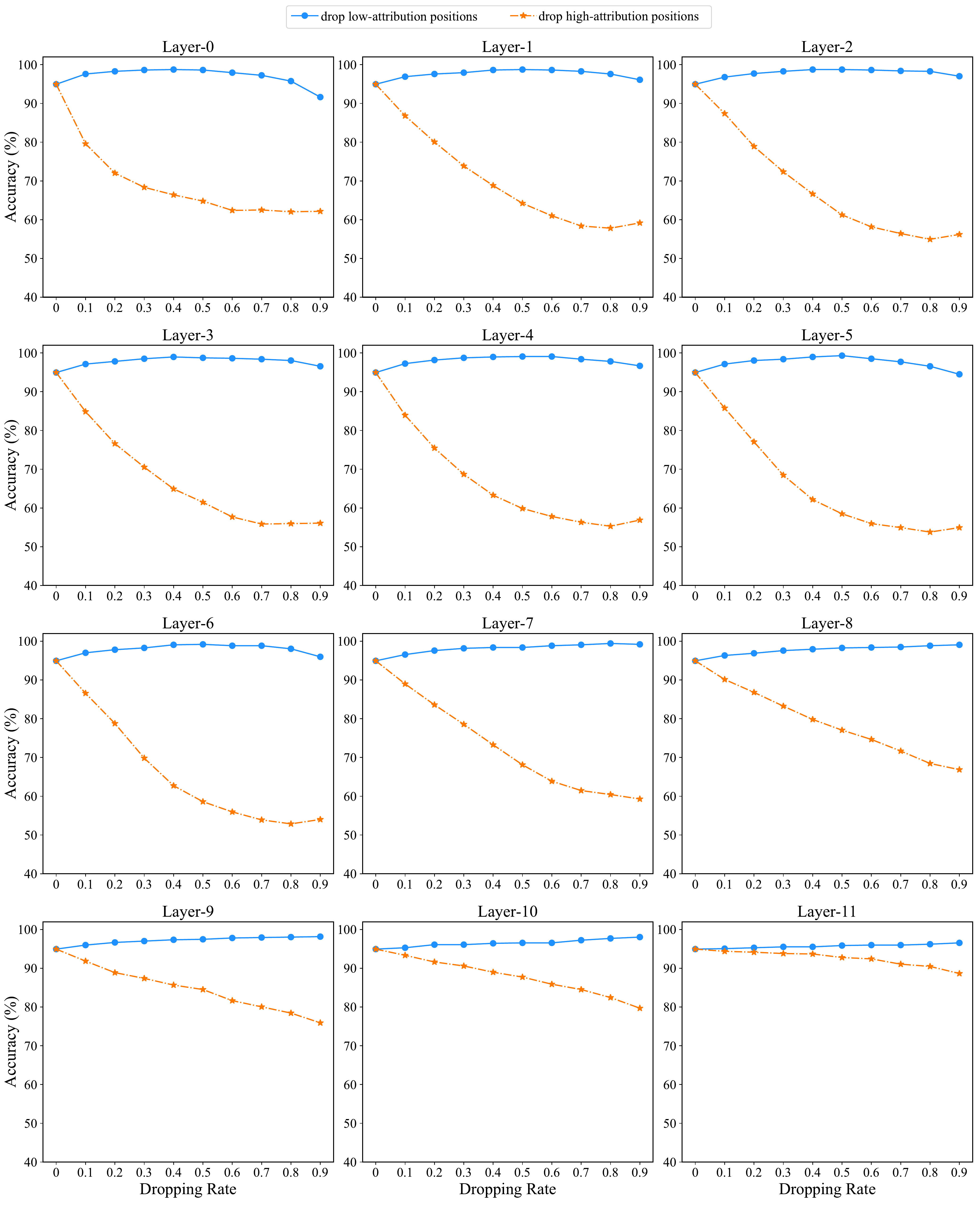}
	\caption{Results of dropping self-attention positions in different layers of RoBERTa on SST-2.}
	\label{fig:appendix_prior_all}
\end{figure*}

Note that we only drop positions in the first layer of RoBERTa for the above experiments to exclude the impact of different layers. We also conduct experiments in the other layers on SST-2, and the overall results are shown in Figure \ref{fig:appendix_prior_all}. We note that similar results are obtained in the first few layers, while the trend becomes less obvious as the number of layers increases. It could be caused by the over-smoothing \cite{shi2021revisiting} issue that the representations of all tokens are similar in the last few layers.

\section{Appendix: Experimental Details}\label{apd:exp}

\begin{wraptable}{r}{6.6cm}
	\centering
	\vspace{-0.5cm}
	\caption{Statistics of the used datasets.}
	\begin{tabular}{lccc}
			\toprule
			Dataset & Train & Dev & Test \\
			\midrule
			SST-2 & 67349 & 872 & 1821  \\
			MNLI & 392702 & 9815 & 9796 \\
			QNLI & 104743 & 5463 & 5463  \\
			QQP & 363846 & 40430 & 390965 \\
			CoLA & 8551 & 1043 & 1063  \\
			STS-B & 5749 & 1500 & 1378  \\
			MRPC & 3668 & 408 & 1725 \\
			RTE & 2490 & 277 & 3000 \\
			CoNLL-2003 & 14041 & 3250 & 3453 \\
			EN-RO & 610320 & 1999 & 1999 \\
			TR-EN & 205756 & 1001 & 3000 \\
			HANS & 30000 & 30000 & - \\
			PAWS-X & 49401 & 2000 & 2000 \\
			\bottomrule
		\end{tabular}
	\label{tab:dataset}
\end{wraptable}

\subsection{Details of Datasets}\label{apd:exp_dataset}
The details of the used datasets are introduced as follows. (1) Stanford Sentiment Treebank (\textbf{SST-2}) \cite{socher2013recursive} is a sentence sentiment prediction task. (2) Multi-Genre Natural Language Inference (\textbf{MNLI}) \cite{williams2017broad} is a pairwise sentence classification task that aims to predict whether the relationship between two sentences is entailment, contradiction, or neutral. (3) Question Natural Language Inference (\textbf{QNLI}) \cite{rajpurkar2016squad} is a binary sentence classification task that aims to predict whether the sentence in a question-sentence pair contains the correct answer to the question. (4) Quora Question Pairs (\textbf{QQP}) \cite{chen2018quora} is a binary pairwise sentence classification task that aims to predict whether two questions are semantically equivalent. (5) The Corpus of Linguistic Acceptability (\textbf{CoLA}) \cite{warstadt2019cola} aims to predict whether a single English sentence conforms to linguistics. (6) The goal of Semantic Textual Similarity Benchmark (\textbf{STS-B}) \cite{cer2017semeval} is to predict how two given sentences are semantically similar. (7) Microsoft Research Paraphrase Corpus (\textbf{MRPC}) \cite{dolan2005automatically} aims to predict if two sentences are semantically equivalent. (8) Recognizing Textual Entailment (\textbf{RTE}) \cite{bentivogli2009fifth} is similar to MNLI but has binary labels. (9) \textbf{CoNLL-2003} \cite{sang2003introduction} is to recognize the named entities in a sentence, which contains four types of named entities. (10) WMT 2016 \cite{bojar2016findings} is a multilingual translation database. In this study, we choose \textbf{English-Romanian (EN-RO)} and \textbf{Turkish-English (TR-EN)} for the experiment. (11) Heuristic Analysis for NLI Systems (\textbf{HANS}) \cite{mccoy2019right} aims to evaluate whether NLI models adopt syntactic heuristics. (12) \textbf{PAWS-X} \cite{yang2019paws} is a cross-lingual adversarial dataset for paraphrase identification. HANS and PAWS-X are typically used for the OOD generalization test. The statistics of these datasets are shown in Table \ref{tab:dataset}. 

\subsection{Hyperparameter Settings}\label{apd:exp_set}

Table \ref{tab:settings} presents the final hyperparameter settings of \textsc{AD-Drop} for BERT/RoBERTa$_{\rm{base}}$. The setting with only one value means the parameter is shared by BERT and RoBERTa.

\begin{table}[htbp]
	\centering
	\caption{Hyperparameter settings of \textsc{AD-Drop} for BERT and RoBERTa.}
	\vspace{0.3cm}
	\resizebox{0.7\textwidth}{!}{
		\begin{tabular}{lcccccc}
			\toprule
			Dataset &  Learning rate & Batch size & Length  & $p$ & $q$\\
			\midrule
			SST-2 &  1e-5 & 16/64 & 120 & 0.6/0.3 & 0.8/0.7 \\
			MNLI &  1e-5 & 16/32 & 128  & 0.5/0.4 & 0.9/0.2\\
			QNLI &  1e-5 & 16 & 128  & 0.8 & 0.8/0.4\\
			QQP &  1e-5 & 16 & 120 & 0.2/0.7 & 0.7/0.9\\
			CoLA &  1e-5/2e-5 & 16 & 47 & 0.3/0.8 & 0.4/0.3 \\
			STS-B & 1e-5/2e-5 & 16 & 100 & 0.9/0.1 & 0.7/0.5 \\
			MRPC & 1e-5/2e-5 & 16 & 100 & 0.5/0.8 & 0.8/0.3\\
			RTE & 1e-5 & 16 & 128 & 0.6/0.7 & 0.7/0.1\\
			\bottomrule
		\end{tabular}
	}
	\label{tab:settings}
\end{table}

\section{Appendix: More Experimental Results}\label{apd:results}

\subsection{Ablation of Cross-Tuning}\label{apd:results_cross}
\begin{figure*}[htbp]
	\centering
	\subfloat[BERT on CoLA]{
		\centering
		\includegraphics[width=0.42\textwidth]{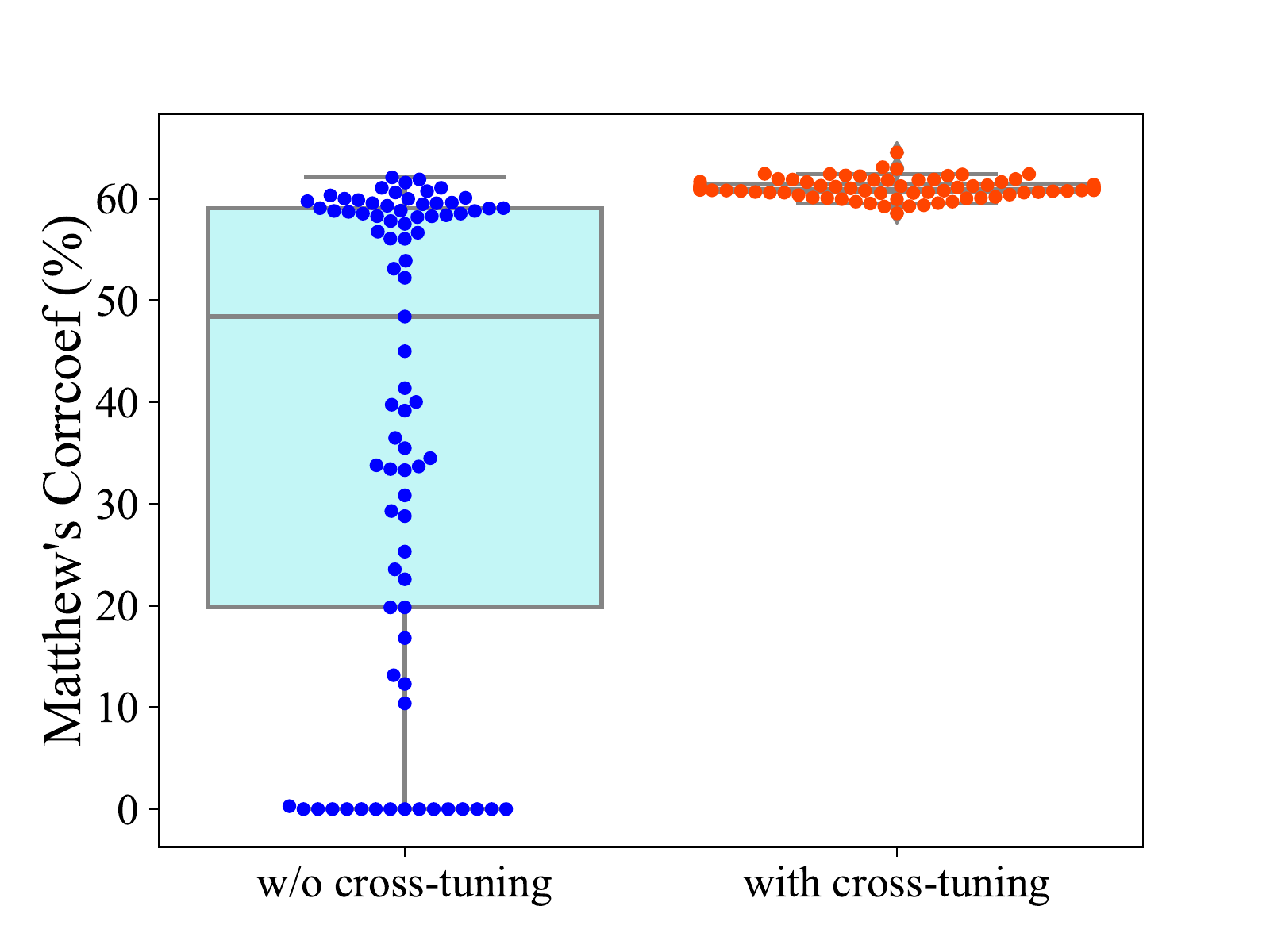}}
		\hspace{5mm}
	\subfloat[RoBERTa on CoLA]{
		\centering
		\includegraphics[width=0.42\textwidth]{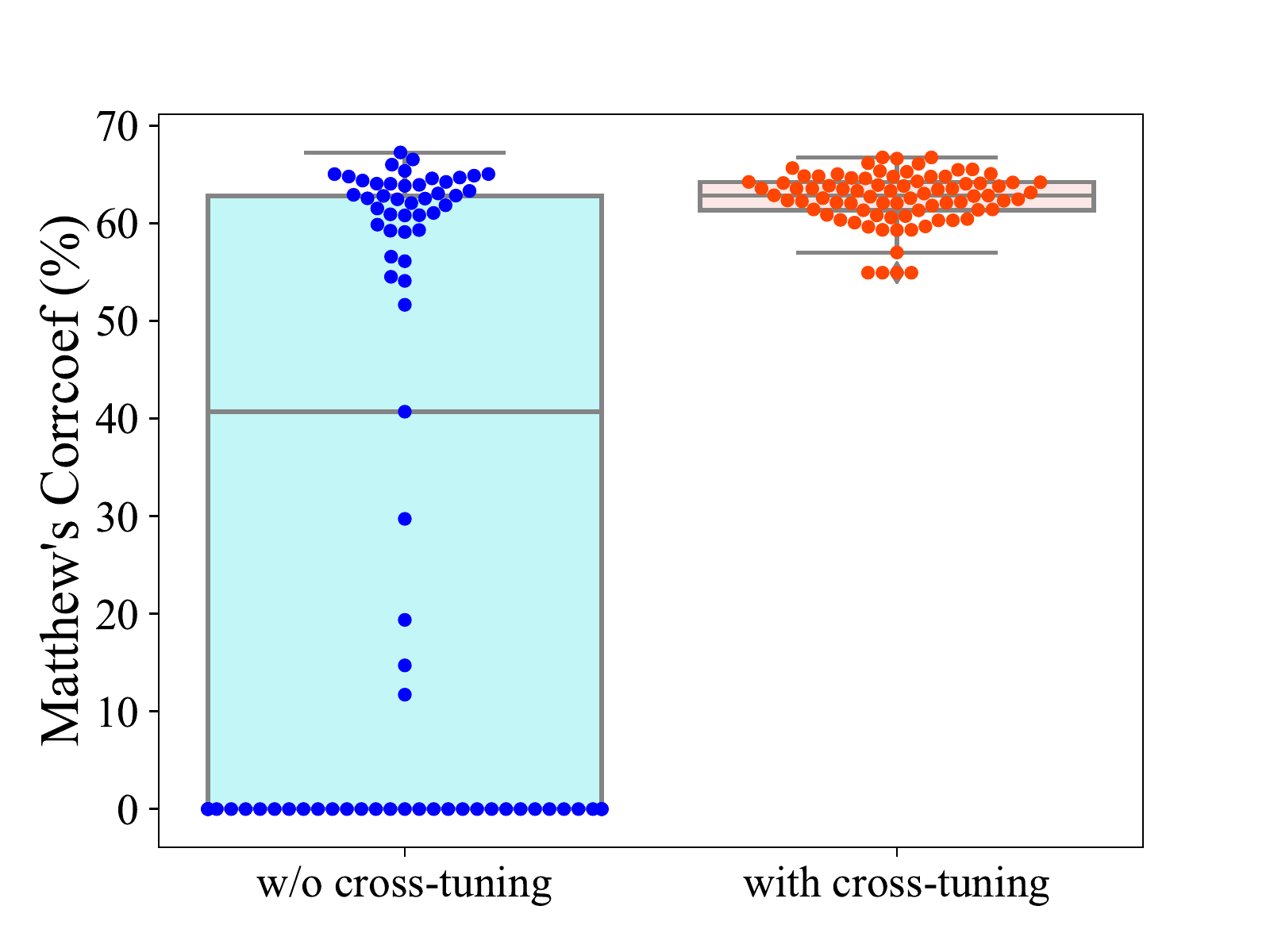}} \\
	\subfloat[BERT on MRPC]{
		\centering
		\includegraphics[width=0.42\textwidth]{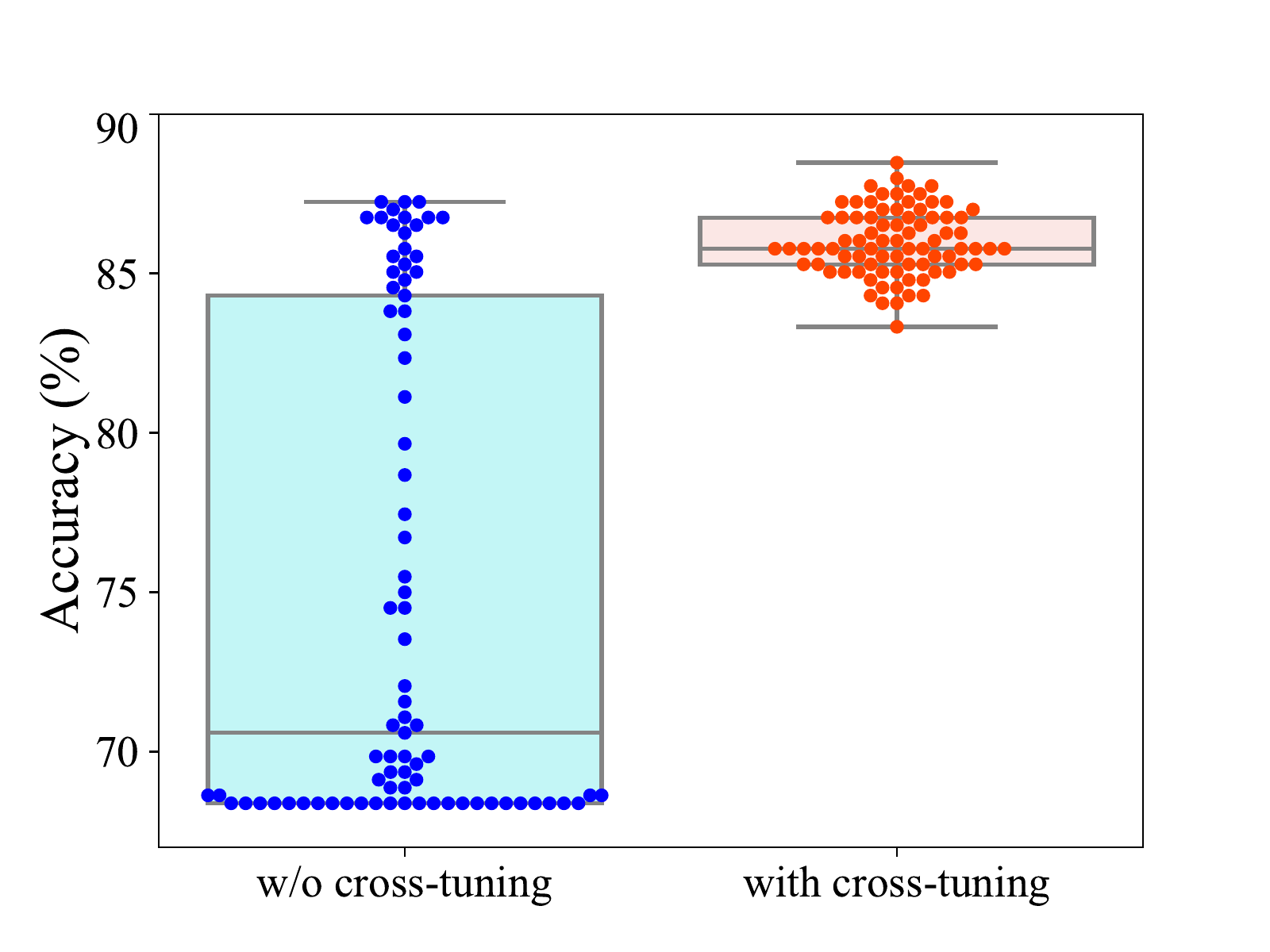}}\hspace{5mm}
	\subfloat[RoBERTa on MRPC]{
		\centering
		\includegraphics[width=0.42\textwidth]{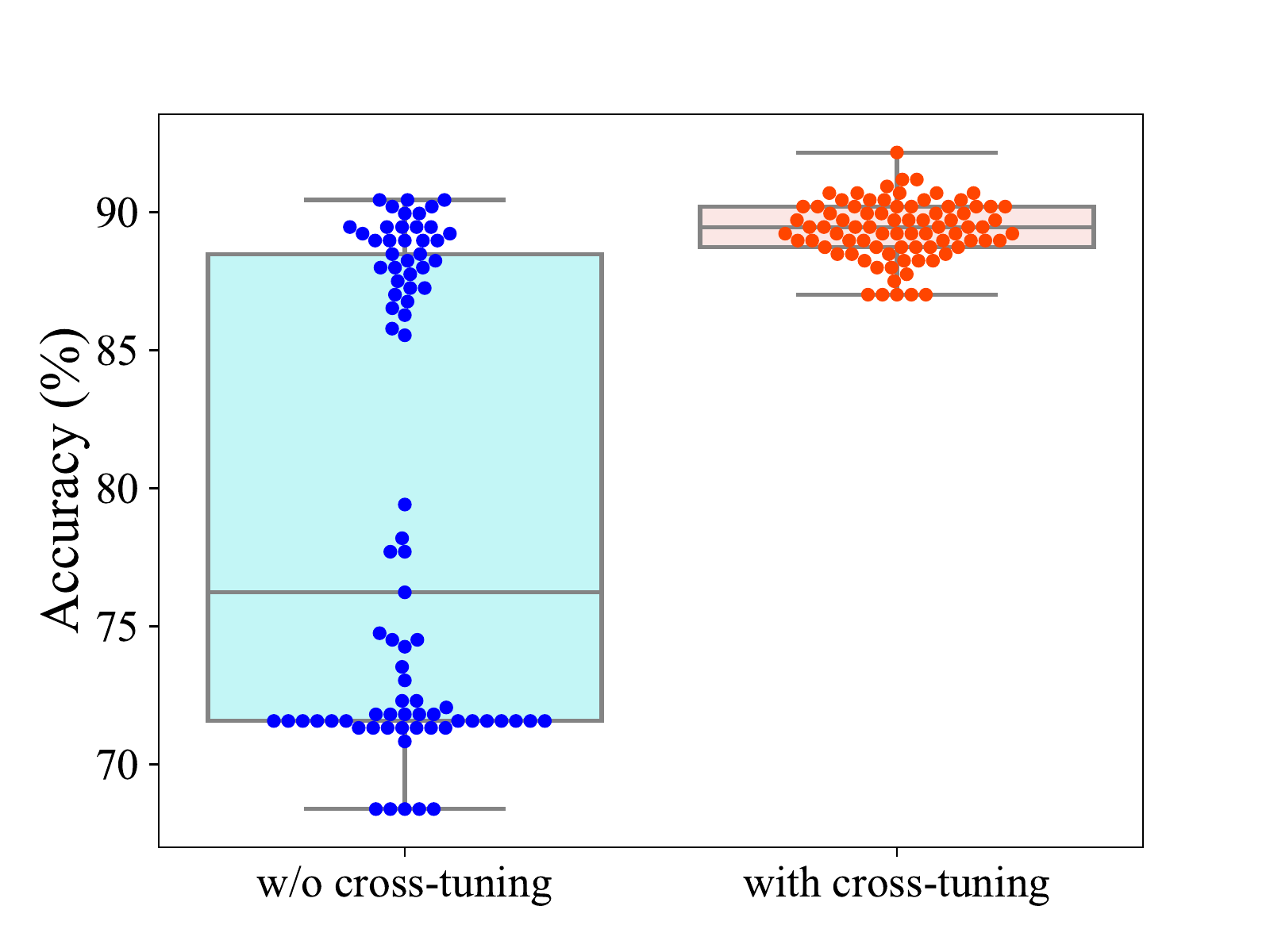}}
	
	\caption{Results of \textsc{AD-Drop} with and without cross-tuning when enumerating $p$ and $q$ in the range of [0.1, 0.9] 	on the CoLA and MRPC datasets.}
	\label{fig:apd_cross}
\end{figure*}

We further report the results of removing cross-tuning in \textsc{AD-Drop} when enumerating $p$ and $q$ in the range of [0.1, 0.9] on the CoLA and MRPC datasets. We observe consistent performance degradation in Figure \ref{fig:apd_cross} after removing the cross-tuning strategy from \textsc{AD-Drop}. 

\begin{wrapfigure}{r}{5.8cm}
	\centering
	\includegraphics[width=0.4\textwidth]{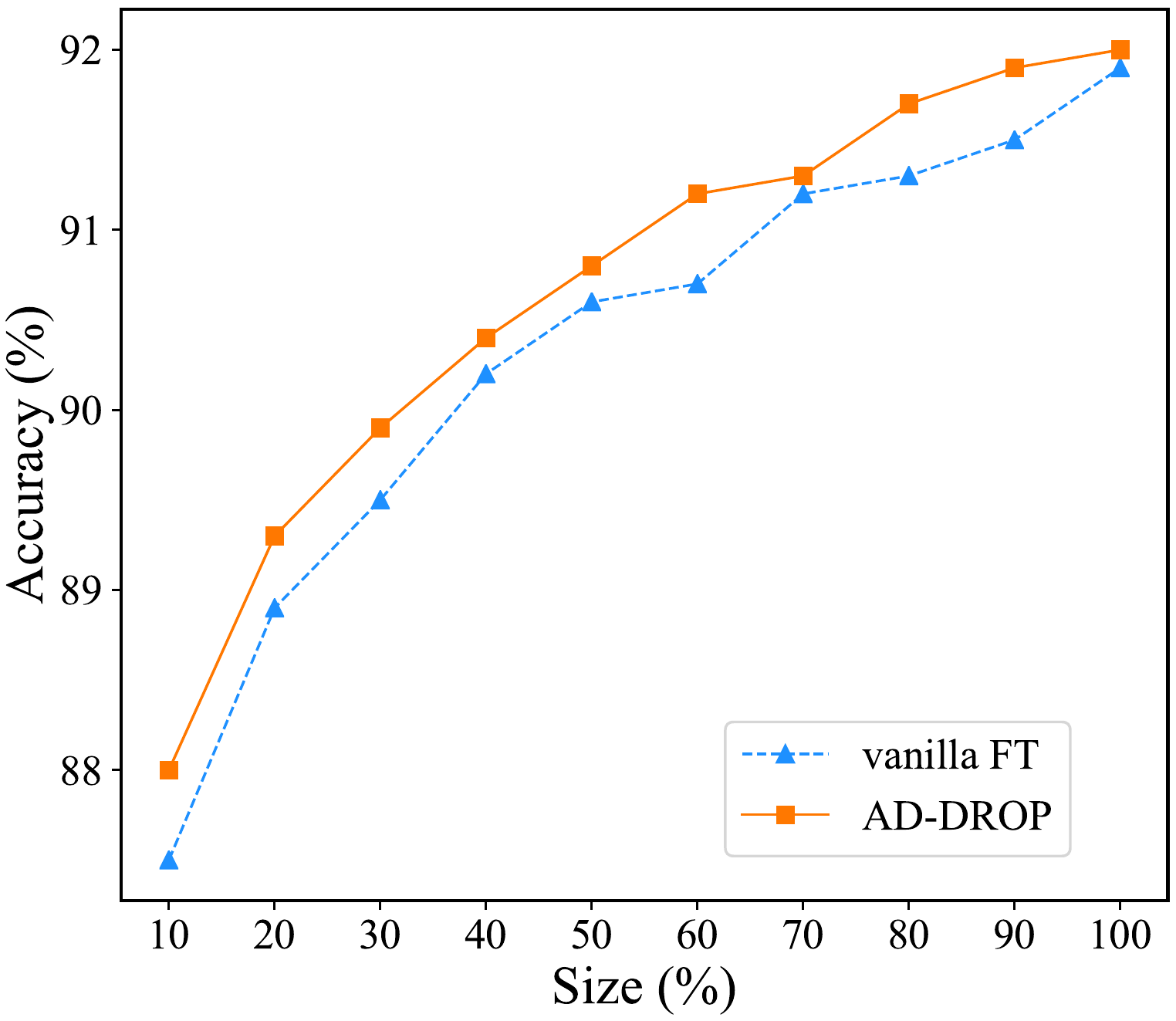}
	\caption{Results of comparison between \textsc{AD-Drop} and original FT as the size of training data changes on QQP.}
	\label{fig:size_qqp}
	\vspace{-1.2cm}
\end{wrapfigure}

\subsection{Effect of Data Size on QQP}\label{apd:qqp}
Figure \ref{fig:size_qqp} shows a comparison between \textsc{AD-Drop} and the original fine-tuning (FT) as the size of training examples changes. We observe from the figure that \textsc{AD-Drop} performs consistently better than original FT with different sizes of training data.

\section{Appendix: Limitations }\label{apd:limit}
We discuss potential limitations of \textsc{AD-Drop} as follows. First, as reported in Section \ref{time}, training with \textsc{AD-Drop} requires more computational cost than the original fine-tuning approach especially when integrated gradient is applied for attribution in all the attention heads. Therefore, we propose to use gradient for attribution in \textsc{AD-Drop} as it achieves competitive performance with acceptable computational cost. Second, \textsc{AD-Drop} introduces additional hyperparameters ($p$ and $q$) and requires more effort to search for the best hyperparameters.
\end{document}